\documentclass[final]{cvpr}

\usepackage{times}
\usepackage{epsfig}
\usepackage{graphicx}
\usepackage{amsmath}
\usepackage{amssymb}

\usepackage{url}            
\usepackage{booktabs}       
\usepackage{amsfonts}       
\usepackage{nicefrac}       
\usepackage{microtype}      
\usepackage{amsmath}
\usepackage{listings}
\usepackage[export]{adjustbox}
\usepackage{caption}
\usepackage{centernot}

\makeatletter
\@namedef{ver@everyshi.sty}{}
\makeatother
\usepackage{tikz}
\usepackage{colortbl}
\usepackage{xspace}

\usepackage[breaklinks=true,bookmarks=false]{hyperref}

\def\cvprPaperID{5246} 
\def\confYear{CVPR 2021}

\usetikzlibrary{positioning}
\usetikzlibrary{shapes.geometric}
\usetikzlibrary{arrows.meta}
\usepackage{pgfplots}
\pgfplotsset{width=7cm,compat=1.8}
\pgfplotsset{colormap={mine}{[1cm] rgb255(0cm)=(255,0,0) rgb255(1cm)=(0,0,255)}}

\author{%
  Eric Heitz 
\hspace{5mm}
  Kenneth Vanhoey
\hspace{5mm}
  Thomas Chambon
\hspace{5mm}
  Laurent Belcour \\
	Unity Technologies \\
  \texttt{\{eric,kennethv,thomas.chambon,laurent\}@unity3d.com} 
}

\graphicspath{{figures/}}

\makeatletter
\def\input@path{{figures/}}
\makeatother

\newcommand{\Gram}{$\mathcal{L}_\text{Gram}$\xspace}
\newcommand{\Hist}{$\mathcal{L}_\text{SW}$\xspace}
\newcommand{\Context}{$\mathcal{L}_\text{CX}$\xspace}
\newcommand{\STROTSSwithreg}{$\mathcal{L}_\text{REMD}$ + $\mathcal{L}_\text{m}$ + $\mathcal{L}_\text{p}$\xspace}
\newcommand{\STROTSSnoreg}{$\mathcal{L}_\text{REMD}$\xspace}
\newcommand{\STROTSS}{$\mathcal{L}_\text{REMD}$\xspace}

\begin{document}

\title{A Sliced Wasserstein Loss for Neural Texture Synthesis}

\maketitle
\thispagestyle{empty}

\begin{abstract}
We address the problem of computing a textural loss based on the statistics extracted from the feature activations of a convolutional neural network optimized for object recognition (\eg VGG-19). 
The underlying mathematical problem is the measure of the distance between two distributions in feature space. 
The Gram-matrix loss is the ubiquitous approximation for this problem but it is subject to several shortcomings.
Our goal is to promote the Sliced Wasserstein Distance as a replacement for it.
It is theoretically proven, practical, simple to implement, and achieves results that are visually superior for texture synthesis by optimization or 
training generative neural networks.
\end{abstract}



\section{Introduction}
\label{sec:intro}

A texture is by definition a class of images that share a set of stationary statistics. 
One of the key components of texture synthesis is a textural loss that measures the difference between two images with respect to these stationary statistics. 

\begin{figure}[t]
    \footnotesize
	\begin{center}
            \begin{tabular}{@{\hspace{-0.2mm}}c@{\hspace{0.5mm}}c@{\hspace{0.5mm}}c@{}}
                \textbf{input}
                & \textbf{optim} {\small \Gram}
                & \textbf{optim} {\small \Hist}
                \\
                \fbox{\includegraphics[width=.3\linewidth]{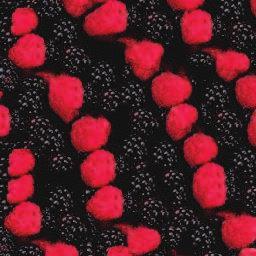}}
                & \fbox{\includegraphics[width=.3\linewidth]{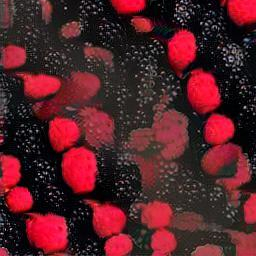}}
                & \fbox{\includegraphics[width=.3\linewidth]{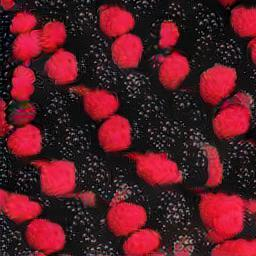}}
                \\
                \scalebox{0.8}{
                \begin{tikzpicture}
                    \draw[-{Triangle[width=6pt,length=2pt]}, line width=3pt](0.5,0.2) -- (0.5, -0.1);
                    \input{vgg_1}
                \end{tikzpicture}
                }
                &
                \scalebox{0.8}{
                \begin{tikzpicture}
                    \draw[-{Triangle[width=6pt,length=2pt]}, line width=3pt](0.5,0.2) -- (0.5, -0.1);
                    \input{vgg_2}
                \end{tikzpicture}
                }
                &
                \scalebox{0.8}{
                \begin{tikzpicture}
                    \draw[-{Triangle[width=6pt,length=2pt]}, line width=3pt](0.5,0.2) -- (0.5, -0.1);
                    \input{vgg_3}
                \end{tikzpicture}
                }
                \\
                \scalebox{1.25}{
                    \begin{tikzpicture}
                        \input{figure1_image.tex}
                    \end{tikzpicture}
                }
                \hspace{-2mm}
                \scalebox{0.95}{
                    \begin{tikzpicture}
                        \begin{scope}[scale=0.185]
\begin{axis}[width=0.5\textwidth,height=0.5\textwidth,grid=major,xmin=0.0,xmax=1.0,ymin=0.0,ymax=1.0,ticks=none]
	\addplot[only marks, scatter,colormap name=mine,point meta=y-x,domain=0:1] 
		coordinates {
( 0.7375242851479528 , 0.07149719324412154 )
( 0.6352616126167896 , 0.3741453717776989 )
( 0.6659961465969623 , 0.16905674183969271 )
( 0.7581160815510639 , 0.1634410081431301 )
( 0.8163931968005627 , 0.14827914918912766 )
( 0.235522015159636 , 0.862679580008819 )
( 0.4358410694203565 , 0.7349179605485335 )
( 0.09431609762856191 , 0.808318117172903 )
( 0.1154706312820831 , 0.7728239727237151 )
( 0.3139804373163412 , 0.9008613797149576 )
( 0.157427062376505 , 0.9106254691891366 )
( 0.8087453710717747 , 0.0873358066465128 )
( 0.22286168569889056 , 0.6672178044029601 )
( 0.22297990357259065 , 0.6156022713570786 )
( 0.8516233378780766 , 0.11998168903668985 )
( 0.19926435538276854 , 0.6994255744573358 )
( 0.38176637786578715 , 0.7412614969652668 )
( 0.12451618465828287 , 0.7369979876185243 )
( 0.10979748007965975 , 0.8512815509980239 )
( 0.10479225209789506 , 0.6579663171400711 )
( 0.7009834334711795 , 0.3720120811638443 )
( 0.7773961056233383 , 0.3599495053784583 )
( 0.8201340945592945 , 0.22151647062355037 )
( 0.131654833648978 , 0.816398412146849 )
( 0.6245513907080336 , 0.13009483281432893 )
( 0.430694727643888 , 0.7696278088773739 )
( 0.1972961888949818 , 0.7946443747181786 )
( 0.8941919835488764 , 0.3450755917066442 )
( 0.5897943690388352 , 0.26685768854268516 )
( 0.8482706859776603 , 0.404644785626021 )
( 0.4284911662411657 , 0.8111718300538617 )
( 0.4076415266589243 , 0.6543043974677796 )
( 0.16102372309131766 , 0.8368607468802273 )
( 0.2701167616942569 , 0.8142574315792738 )
( 0.1539149012676253 , 0.6483720921974706 )
( 0.2841292635125761 , 0.7305270744563124 )
( 0.9277902148119016 , 0.1938935051088681 )
( 0.27391219388005256 , 0.5725148987449746 )
( 0.36551326907490445 , 0.7713106951700408 )
( 0.0734335217243964 , 0.7396019784732524 )
( 0.6543978342157951 , 0.08650271025822998 )
( 0.6512898381036001 , 0.2000216613180242 )
( 0.1995156704890989 , 0.8120878635117923 )
( 0.821622311898793 , 0.2790205918520516 )
( 0.6962408637146551 , 0.08076108729329323 )
( 0.3636534200400908 , 0.8083595677085642 )
( 0.2679208190322526 , 0.6244303950353428 )
( 0.10389673801250013 , 0.8681040527778956 )
( 0.8033544609775752 , 0.36867725590955347 )
( 0.2740573751979177 , 0.7243578756697717 )
( 0.6948557533715101 , 0.18925959140395698 )
( 0.6728557525654665 , 0.2679928808749045 )
( 0.8647811959996955 , 0.18825570536856379 )
( 0.23518869693404476 , 0.9128898340829545 )
( 0.265926429611197 , 0.9133710202209234 )
( 0.6220106906020064 , 0.1219079342588435 )
( 0.6902126391538932 , 0.23480703283724697 )
( 0.08131589045249188 , 0.6954331462900474 )
( 0.7832485755015384 , 0.2709581697935482 )
( 0.19700724066228567 , 0.9067820388330438 )
( 0.7340980139455315 , 0.2627891475368584 )
( 0.22544026580990742 , 0.806029256851962 )
( 0.6648862020343914 , 0.3593651794811017 )
( 0.20270234187784264 , 0.640557176544345 )
};
	\end{axis}
\end{scope}
                    \end{tikzpicture}
                }
                &
                \scalebox{1.25}{
                    \begin{tikzpicture}
                        \input{figure2_image.tex}
                    \end{tikzpicture}
                }
                \hspace{-2mm}
                \scalebox{0.95}{
                    \begin{tikzpicture}
                        \begin{scope}[scale=0.185]
\begin{axis}[width=0.5\textwidth,height=0.5\textwidth,grid=major,xmin=0.0,xmax=1.0,ymin=0.0,ymax=1.0,ticks=none]
	\addplot[only marks, scatter,colormap name=mine,point meta=y-x,domain=0:1] 
		coordinates {
( 0.8202041704413296 , 0.07628025353023737 )
( 0.4984247674129366 , 0.18288786191229237 )
( 0.9268010853831867 , 0.09608852539259066 )
( 0.5273644078112081 , 0.39197019254644644 )
( 0.4194488001063227 , 0.8261326504107683 )
( 0.5743722550301275 , 0.696873432205544 )
( 0.9289205660903701 , 0.27286030753684837 )
( 0.2007151986453794 , 0.6598150333145507 )
( 0.46224454418719985 , 0.7329328567767481 )
( 0.342876958634608 , 0.4017393904958452 )
( 0.3543545048940159 , 0.9422365848723856 )
( 0.6466234862868944 , 0.6416022761181387 )
( 0.15648664390872535 , 0.7180661420327928 )
( 0.6223219967194824 , 0.3482228422560979 )
( 0.48828781340915783 , 0.32001741705530184 )
( 0.12761947313259792 , 0.9497758907046843 )
( 0.7188779739238289 , 0.31172419194823897 )
( 0.0756814074676318 , 0.9205740450269606 )
( 0.928600733336744 , 0.1522930123391687 )
( 0.30548567287672435 , 0.498694337397388 )
( 0.7543365439343221 , 0.5197342844826867 )
( 0.5906880260232503 , 0.7574932299904413 )
( 0.7316858778427182 , 0.1631244864843146 )
( 0.7384460460186435 , 0.19313601268748143 )
( 0.24242220976055634 , 0.940164811579578 )
( 0.6097223989603282 , 0.09038082550318216 )
( 0.4131667928355628 , 0.6238257908972612 )
( 0.25715326144486506 , 0.3821452501067999 )
( 0.6962678910242377 , 0.5774136803984509 )
( 0.3491204393574636 , 0.5547379482787673 )
( 0.4942989078304718 , 0.46653182517714653 )
( 0.05955195873105679 , 0.9936523419991984 )
( 0.07034584632316154 , 0.7277246903958884 )
( 0.7285611159686596 , 0.027914974447377894 )
( 0.46453807952613024 , 0.2680759846302929 )
( 0.8926840043927275 , 0.4276633955525242 )
( 0.920710421618857 , 0.02324161392944324 )
( 0.7499289253737385 , 0.09513713935319129 )
( 0.2684421676029399 , 0.8037268753642233 )
( 0.6737868847086173 , 0.24543000794190203 )
( 0.6346858784941599 , 0.5172903443628526 )
( 0.572328152021811 , 0.7769033116284686 )
( 0.6797034640804087 , 0.036657815009165595 )
( 0.17886925902107464 , 0.7717353371789318 )
( 0.8151148438886053 , 0.25122374955468185 )
( 0.3807214126354659 , 0.2777372143317143 )
( 0.5647717477877299 , 0.5715069340329069 )
( 0.7174546089773177 , 0.3995517539371371 )
( 0.3819976728982321 , 0.42978443837280916 )
( 0.44928694175315464 , 0.6447437852375126 )
( 0.592360810638002 , 0.3972306209245981 )
( 0.6856733656422312 , 0.1289147103298487 )
( 0.08879323152023322 , 0.5817559945870591 )
( 0.3307323145838629 , 0.727118574122373 )
( 0.08355242309610672 , 0.6591022010229066 )
( 0.4564596408612209 , 0.6066385493894038 )
( 0.8472730866087572 , 0.3444127526442948 )
( 0.05876583812587921 , 0.8045299232976187 )
( 0.8117260943758872 , 0.3218448607194123 )
( 0.15932066528061012 , 0.6446497372804534 )
( 0.24413452239929478 , 0.8515490427939978 )
( 0.31702340827098424 , 0.6153743276293124 )
( 0.7381643489023157 , 0.4023840274749544 )
( 0.15168443317060087 , 0.5079963443622656 )
};
    \draw[dashed,gray] \pgfextra{
	  \pgfpathellipse{\pgfplotspointaxisxy{0.5}{0.5}}
		{\pgfplotspointaxisdirectionxy{0.5}{-0.5}}
		{\pgfplotspointaxisdirectionxy{0.2}{0.2}}
	};

	\end{axis}
\end{scope}
                    \end{tikzpicture}
                }
                &
                \scalebox{1.25}{
                    \begin{tikzpicture}
                        \input{figure3_image.tex}
                    \end{tikzpicture}
                }
                \hspace{-2mm}
                \scalebox{0.95}{
                    \begin{tikzpicture}
                        \begin{scope}[scale=0.185]
\begin{axis}[width=0.5\textwidth,height=0.5\textwidth,grid=major,xmin=0.0,xmax=1.0,ymin=0.0,ymax=1.0,ticks=none]
	\addplot[only marks, scatter,colormap name=mine,point meta=y-x,domain=0:1] 
		coordinates {
( 0.7375242851479528 , 0.07149719324412154 )
( 0.6352616126167896 , 0.3741453717776989 )
( 0.6659961465969623 , 0.16905674183969271 )
( 0.7581160815510639 , 0.1634410081431301 )
( 0.8163931968005627 , 0.14827914918912766 )
( 0.235522015159636 , 0.862679580008819 )
( 0.4358410694203565 , 0.7349179605485335 )
( 0.09431609762856191 , 0.808318117172903 )
( 0.1154706312820831 , 0.7728239727237151 )
( 0.3139804373163412 , 0.9008613797149576 )
( 0.157427062376505 , 0.9106254691891366 )
( 0.8087453710717747 , 0.0873358066465128 )
( 0.22286168569889056 , 0.6672178044029601 )
( 0.22297990357259065 , 0.6156022713570786 )
( 0.8516233378780766 , 0.11998168903668985 )
( 0.19926435538276854 , 0.6994255744573358 )
( 0.38176637786578715 , 0.7412614969652668 )
( 0.12451618465828287 , 0.7369979876185243 )
( 0.10979748007965975 , 0.8512815509980239 )
( 0.10479225209789506 , 0.6579663171400711 )
( 0.7009834334711795 , 0.3720120811638443 )
( 0.7773961056233383 , 0.3599495053784583 )
( 0.8201340945592945 , 0.22151647062355037 )
( 0.131654833648978 , 0.816398412146849 )
( 0.6245513907080336 , 0.13009483281432893 )
( 0.430694727643888 , 0.7696278088773739 )
( 0.1972961888949818 , 0.7946443747181786 )
( 0.8941919835488764 , 0.3450755917066442 )
( 0.5897943690388352 , 0.26685768854268516 )
( 0.8482706859776603 , 0.404644785626021 )
( 0.4284911662411657 , 0.8111718300538617 )
( 0.4076415266589243 , 0.6543043974677796 )
( 0.16102372309131766 , 0.8368607468802273 )
( 0.2701167616942569 , 0.8142574315792738 )
( 0.1539149012676253 , 0.6483720921974706 )
( 0.2841292635125761 , 0.7305270744563124 )
( 0.9277902148119016 , 0.1938935051088681 )
( 0.27391219388005256 , 0.5725148987449746 )
( 0.36551326907490445 , 0.7713106951700408 )
( 0.0734335217243964 , 0.7396019784732524 )
( 0.6543978342157951 , 0.08650271025822998 )
( 0.6512898381036001 , 0.2000216613180242 )
( 0.1995156704890989 , 0.8120878635117923 )
( 0.821622311898793 , 0.2790205918520516 )
( 0.6962408637146551 , 0.08076108729329323 )
( 0.3636534200400908 , 0.8083595677085642 )
( 0.2679208190322526 , 0.6244303950353428 )
( 0.10389673801250013 , 0.8681040527778956 )
( 0.8033544609775752 , 0.36867725590955347 )
( 0.2740573751979177 , 0.7243578756697717 )
( 0.6948557533715101 , 0.18925959140395698 )
( 0.6728557525654665 , 0.2679928808749045 )
( 0.8647811959996955 , 0.18825570536856379 )
( 0.23518869693404476 , 0.9128898340829545 )
( 0.265926429611197 , 0.9133710202209234 )
( 0.6220106906020064 , 0.1219079342588435 )
( 0.6902126391538932 , 0.23480703283724697 )
( 0.08131589045249188 , 0.6954331462900474 )
( 0.7832485755015384 , 0.2709581697935482 )
( 0.19700724066228567 , 0.9067820388330438 )
( 0.7340980139455315 , 0.2627891475368584 )
( 0.22544026580990742 , 0.806029256851962 )
( 0.6648862020343914 , 0.3593651794811017 )
( 0.20270234187784264 , 0.640557176544345 )
};
	\end{axis}
\end{scope}
                    \end{tikzpicture}
                }
            \end{tabular}
	\end{center}
\vspace{-6mm}
 \caption{\textbf{Neural texture synthesis.} We visualize a 2D slice of the feature distributions of a convolutional layer. 
The ubiquitous Gram-Matrix loss \Gram captures only the major directions of the distribution.
We promote the Sliced Wasserstein Distance \Hist that captures the full distribution and allows for neural texture synthesis with improved quality.}
\label{fig:teaser}
\vspace{-5mm}
\end{figure}

Gatys~\etal~\cite{Gatys15} discovered that the feature activations in pretrained Convolutional Neural Networks (CNNs) such as VGG-19~\cite{Simonyan14} yield powerful textural statistics.
Neural texture synthesis means optimizing an image to match the feature distributions of a target texture in each convolutional layer, as shown in Figure~\ref{fig:teaser}.
Gatys~\etal use the $L^2$ distance between the Gram matrices of the feature distributions as a textural loss \Gram. 
The simplicity and practicability of this loss make it especially attractive and it is nowadays ubiquitous in neural texture synthesis and style 
transfer 
methods~\cite{Gatys16,Gatys16color,UlyanovV1,SpectrumConstraints,UlyanovVL17,DiversifiedTexSynth,MRTexSynth,DeepCorrelation,NonStationaryTexSynth,
TextureMixer19}.

\setlength{\fboxsep}{0pt}\setlength{\fboxrule}{0.8pt}

Intuitively, the Gram-matrix loss \Gram, which is a second-order descriptor like the covariance matrix, optimizes the features to be distributed along the same major directions but misses other (\eg higher-order) statistics and is thus insufficient to represent the precise shape of the distribution.
This results in undesired artifacts in the synthesized results such as contrast oscillation, as shown in Figure~\ref{fig:teaser}-middle.
Several subsequent works hint that improved textural quality can be obtained by capturing more statistics~\cite{SpectrumConstraints,Risser17,ContextualLoss}. 
However, how to define a practical textural loss that captures the \emph{complete} feature distributions remains an open problem. 

Our main point is that the classic color transfer algorithm of Pitie~\etal~\cite{Pitie05} provides a simple solution to this problem.
This concept is also known as the \emph{Sliced Wasserstein Distance} (SWD)~\cite{rabin2012,bonneel2015}.
It is a slicing algorithm that matches arbitrary n-dimensional (nD) distributions, which is precisely the textural loss problem that the neural texture synthesis community tries to solve. 
Although it has been around and well-studied for a long time, it surprisingly has not yet been considered for neural texture synthesis and we wish to promote it in this context. 
In Section~\ref{sec:contrib}, we show that the SWD can be transposed to deep feature spaces to obtain a textural loss \Hist that captures the complete feature distributions and that is practical enough to be considered as a replacement for \Gram.
Furthermore, we show in Section~\ref{sec:spatial} how to account for user-defined spatial constraints without changing \Hist and without adding further losses or fine-tuned parameters, extending the range of applications reachable with a single textural loss without added complexity.


\section{Problem Statement}
\label{sec:problem_statement}

Our objective is to define a textural loss that captures the complete feature distributions of $L$ target layers of a pretrained convolutional neural network.
For our experiments, we use the first $L=12$ convolutional layers of a pretrained VGG-19 with normalized weights, following Gatys~\etal~\cite{Gatys15}.

\paragraph{Notations.}

The convolutional layer $l$ has $M_l$ pixels (spatial dimensions) and $N_l$ features (depth dimension).
We note $F^l_{m} \in \mathbb{R}^{N_l}$ the feature vector located at pixel $m$ and $F^l_{m}[n] \in \mathbb{R}$ its n-th component ($n < N_l)$.

\paragraph{Deep feature distributions.}

We note $p_l$ the probability density function of the features in layer $l$. Since the feature activations are discrete in a convolutional neural network, the density is a sum of delta Dirac distributions
\begin{align}
p^l(x) = \frac{1}{M_l} \,\sum_{m=1}^{M_l} \delta_{F^l_{m}}(x)
\end{align}
that can be visualized as a point cloud in feature space (Figure~\ref{fig:teaser}).
These distributions are position-agnostic, they do not depend on where the features are located in image space, and provide a stationary statistic of the texture. 

\paragraph{Textural loss.}

A textural loss between two images $I$ and $\tilde I$ is a function $\mathcal{L}(I, \tilde I) \in \mathbb{R}^+$ that measures a distance between the sets of distributions $p^l$ and $\tilde p^l$ associated with the images. 

\paragraph{Objective.}

Our goal is to define a textural loss that captures full feature distributions, \ie
\begin{align}
\label{eq:problem_implication}
\mathcal{L}(I, \tilde I) = 0 \implies p^l = \tilde p^l~~\forall~l \in \{1, .., L\}.
\end{align}
Furthermore, this loss should be practical enough to be used for texture optimization or training generative networks.


\section{Previous work}
\label{sec:previous}

\paragraph{The Gram loss.}

Gatys~\etal~\cite{Gatys15,Gatys16} use the Gram matrices of the feature distributions to define a textural loss:
\begin{align}
\mathcal{L}_{\tiny\text{Gram}}(I, \tilde I)  &= \sum_{l=1}^L \frac{1}{N^2_l} \, \left\|G^l - \tilde G^l  \right\|^2,
\end{align}
where $G^l$ (resp. $\tilde G^l$) is the Gram matrix of the deep features extracted from $I$ (resp. $\tilde I$) at layer $l$.
$G_{ij}^l $ is the entry $(i,j)$ of the Gram matrix $G^l \in \mathbb{R}^{N_l \times N_l}$ of layer $l$, defined as the second-order cross-moment of features $i$ and $j$ over the pixels:
\begin{align}
G_{ij}^l &= \mathbb{E}\left[F^l_{m}[i]  \, F^l_{m}[j]\right] = \frac{1}{M^l} \, \sum_{m} F^l_{m}[i]  \, F^l_{m}[j].
\end{align}
The Gram loss has become ubiquitous as a textural loss because it is fast to compute and practical. 
However, it does not capture the full distribution of features, \ie 
\begin{align}
\label{eq:gram_implication}
\mathcal{L}_{\tiny\text{Gram}}(I, \tilde I) = 0 \centernot\implies p^l = \tilde p^l~~\forall~l \in \{1, .., L\}.
\end{align}
This explains the visual artifacts of Figure~\ref{fig:teaser}-middle.

\paragraph{Beyond the Gram loss.}

Many have noticed that \Gram does not capture every aspect of appearance, resulting in artifacts~\cite{SpectrumConstraints,LiWCT17,Risser17,MRTexSynth,DeepCorrelation,ContextualLoss,NonStationaryTexSynth,TextureMixer19}.
Some approaches switch paradigm by training Generative Adverserial Networks (GANs) but this is out of the scope of the problem defined in Section~\ref{sec:problem_statement} that is the focus of this article.  

The closest approach to ours is the one of Risser~\etal~\cite{Risser17} who define a histogram loss $\mathcal{L}_{\tiny\text{Hist}}$ by adding the sum of the 1D axis-aligned histogram losses of each of the $N_l$ features to the Gram loss:
\begin{align}
\hspace{-2mm}
\mathcal{L}_{\tiny\text{Hist}}(I, \tilde I) = \sum_{l=1}^{L} \sum_{n=1}^{N_l} \mathcal{L}_{\tiny\text{Hist1D}}(p^l_n, \tilde p^l_n) + \alpha \, \mathcal{L}_{\tiny\text{Gram}}(I, \tilde I).
\end{align}
This loss does not capture all the stationary statistics:
\begin{align}
\label{eq:risser_implication}
\mathcal{L}_{\tiny\text{Hist}}(I, \tilde I) = 0 \centernot\implies p^l = \tilde p^l~~\forall~l \in \{1, .., L\},
\end{align}
but significantly improves the results in comparison to \Gram, hinting that capturing more statistics improves textural quality.  
In terms of practicability, a careful tuning of its relative weight is required. 
Moreover, each 1D histogram loss $\mathcal{L}_{\tiny\text{Hist1D}}$ uses a histogram binning.
The number of bins is yet another sensitive parameter to tune: insufficient bins result in poor accuracy while the opposite results in vanishing gradient problems.
Our Sliced Wasserstein loss also computes 1D losses but with an optimal transport formulation (implemented by a sort) rather than a binning scheme and with arbitrary rather than axis-aligned directions, which makes it statistically complete, simpler to use, and allows to get rid of the Gram loss and its weighting parameter.

Other approaches were designed inspired by the idea of capturing the full distribution of features.  
Mechrez~\etal's contextual loss \Context ~\cite{ContextualLoss} estimates the difference between two distributions and resembles Li and Malik's implicit maximum likelihood estimator~\cite{IMLE18}.
Unfortunately, it uses a kernel with an extra meta-parameter~$h$ to tune and has a quadratic complexity, severely limiting its usability.
Another recent work is Kolkin~\etal's approximate earth mover's distance loss \STROTSS~\cite{Kolkin_2019_CVPR}. 
It requires several regularization terms with fine-tuned weights.
These losses are still not complete:
\begin{align}
\label{eq:context_strotss_implication}
\mathcal{L}_{\tiny\text{CX}}(I, \tilde I) = 0 \centernot\implies p^l = \tilde p^l~~\forall~l \in \{1, .., L\}, \\
\mathcal{L}_{\tiny\text{REMD}}(I, \tilde I) = 0 \centernot\implies p^l = \tilde p^l~~\forall~l \in \{1, .., L\},
\end{align}
and are not good candidates for the texture synthesis application for which they achieve textural quality even inferior to a vanilla \Gram.
We further discuss the problems arising when using \Context and \STROTSS for texture synthesis in Appendix~\ref{sec:appendix}.
To our knowledge, a loss that is proven to capture the full distribution of features and provides a practical candidate for texture synthesis has not been proposed yet. 
We believe that the Sliced Wasserstein Distance is the right candidate for this problem.

\paragraph{The Sliced Wasserstein Distance.}

Our main source of inspiration is the color transfer algorithm of Pitie~\etal~\cite{Pitie05}.
They show that iteratively matching random 1D marginals of an n-Dimensional distribution is a sufficient condition to converge towards the distribution, \ie satisfying Equation~(\ref{eq:problem_implication}).
This idea has been applied in the context of optimal transport~\cite{rabin2012,bonneel2015}: distances between distributions can be measured with the Wasserstein Distance and the expectation over random 1D marginals provides a practical approximation that scales in $\mathcal{O}(n\log{}n)$. 
Note that it is approximate in the sense that the optimized transport map is not optimal but the optimized distribution is proven to converge towards the target distribution, \ie it satisfies Equation~(\ref{eq:problem_implication}).
The Sliced Wasserstein Distance allows for fast gradient descent algorithms and is suitable for training generative neural networks~\cite{Kolouri18,Deshpande18,wu19}.
It has also been successfully used for texture synthesis by gradient descent using wavelets as a feature extractor~\cite{Tartavel2016}. 
By using a pretrained CNN as a feature extractor~\cite{Gatys15}, we bring this proven and practical solution to neural texture synthesis and solve the problem defined in Section~\ref{sec:problem_statement}.

\paragraph{Neural texture synthesis with spatial constraints.}

The problem defined in Section~\ref{sec:problem_statement} aims at capturing the stationary statistics that define a texture.
For some applications, additional non-stationary spatial constraints are required.
The mainstream way to handle spatial constraints in neural texture synthesis is to combine \Gram with additional losses~\cite{Champandard16,SpectrumConstraints,DeepCorrelation,NonStationaryTexSynth}. 
For example: Liu~\etal add a loss on the power spectrum of the texture so as to preserve the frequency information~\cite{SpectrumConstraints};
Sendik and Cohen-Or~\cite{DeepCorrelation} add a loss function capturing deep feature correlations with shifted versions of themselves, which makes the optimisation at least an order of magnitude slower. 
All those losses require tuning of both inter-loss weights and inter-layer weights within each loss.
Champandard~\cite{Champandard16} proposes to concatenate user-defined guidance maps to the deep features prior to extracting statistics.
Gatys~\etal~\cite{Gatys_2017_CVPR} show that this hinders textural quality and propose a more qualitative but slower per-cluster variant, which duplicates computation per tag, disallowing numerous tags.
Both methods have sensitive parameters to fine-tune.
In Section~\ref{sec:spatial}, we show how to handle user-defined spatial constraints without modifying the Sliced Wasserstein loss nor adding further 
losses or compromising complexity.


\section{The Sliced Wasserstein Loss}
\label{sec:contrib}

In this section, we show how to compute a neural loss with the Sliced Wasserstein Distance~\cite{Pitie05,Kolouri18,Deshpande18,wu19} and we note it $\mathcal{L}_{\tiny\text{SW}}$.

\begin{figure}[h]
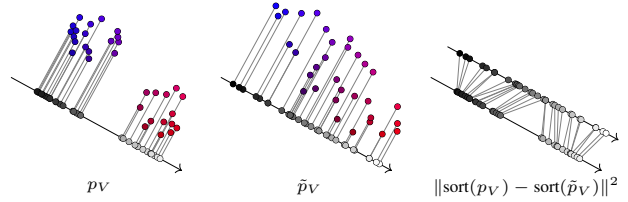

\begin{center}
\begin{tabular}{@{\hspace{0mm}} c @{}}
        \scalebox{0.93}{
        \begin{tikzpicture}
        \begin{scope}[scale=2.0]
        \draw[anchor=center] (0.3, -0.3) node {\scriptsize $p_V$};
        \input{figure1_points_projection.tex}
        \end{scope}
        
        \begin{scope}[shift={(3.0,0.0)}, scale=2.0]
        \draw[anchor=center] (0.3, -0.3) node {\scriptsize $\tilde p_V$};
        \input{figure2_points_projection.tex}
        \end{scope}
        
        \begin{scope}[shift={(6.0,0.0)}, scale=2.0]
        \draw[anchor=center] (0.35, -0.3) node {\scriptsize ${\|\text{sort}(p_V) - \text{sort}(\tilde p_V )}\|^2$};
        \input{figure_loss.tex}
        \end{scope}
        \end{tikzpicture}
        }
\end{tabular}
\end{center}
\vspace{-5mm}
    \caption{\label{fig:sw1dloss} \textbf{The Sliced Wasserstein loss}. 
    We project n-dimensional features onto random directions, sort the 1D projections, and compute the $L^2$ difference between the sorted lists.}
\vspace{-2mm}
\end{figure}

\paragraph{Definition.}

We define the Sliced Wasserstein loss as the sum over the layers:
\begin{equation}
\label{eq:FullLoss}
\mathcal{L}_{\tiny\text{SW}}(I, \tilde I) = \sum_{l=1}^{L} \mathcal{L}_{\tiny\text{SW}}(p^l, \tilde p^l).
\end{equation}
where $\mathcal{L}_{\tiny\text{SW}}(p^l, \tilde p^l)$ is a Sliced Wasserstein Distance between the distribution of features $p^l$ and $\tilde p^l$ of layer $l$. 
It is the expectation of the 1D optimal transport distances after projecting the feature points onto random directions $V \in \mathcal{S}^{N_l}$ on the unit n-dimensional hypersphere of features:
\begin{equation}
\label{eq:histogram_loss}
\mathcal{L}_{\text{SW}}(p_l, \tilde p_l) =  \mathbb{E}_V[
\mathcal{L}_{\tiny\text{SW1D}}(p_V^l, \tilde p_V^l)
],
\end{equation}
where $p_V^l = \{\langle F_m^l,V \rangle\}, \forall m$ is the unordered scalar set of dot products between the $m$ feature vectors $F_m^l$ and the direction $V$. $\mathcal{L}_{\tiny\text{SW1D}}$ is the 1D optimal transport loss between two unordered set of scalars. 
It is defined as the element-wise $L^2$ distance over sorted lists:
\begin{align}
\label{eq:slicing_loss}
    \mathcal{L}_{\text{SW1D}}(S,\tilde S) = \frac{1}{|S|} \, \left\|\text{sort}(S) - \text{sort}(\tilde S)  \right\|^2.
\end{align}
We illustrate the projection, sorting and distance between $p_V^l$ and $\tilde p_V^l$ in Figure~\ref{fig:sw1dloss}.

\paragraph{Properties.}

Pitie~\etal~\cite{Pitie05} prove that the SWD captures the complete target distribution, \ie
\begin{align}
\label{eq:swd_implication1}
\mathcal{L}_{\tiny\text{SW}}(p^l, \tilde p^l) = 0 \implies p^l = \tilde p^l.
\end{align}
It follows that it satisfies the implication of Eq.~(\ref{eq:problem_implication}):
\begin{align}
\label{eq:swd_implication2}
\mathcal{L}_{\tiny\text{SW}}(I, \tilde I) = 0 \implies p^l = \tilde p^l~~\forall~l \in \{1, .., L\},
\end{align}
and hence solves the problem targeted in this paper (Sec.~\ref{sec:problem_statement}).

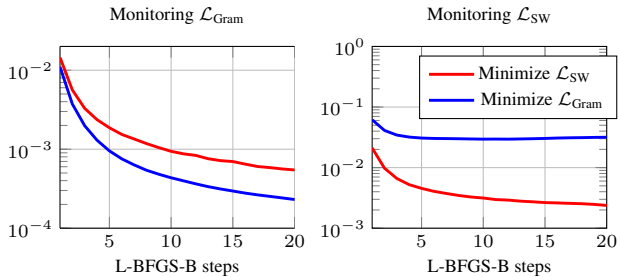
\begin{figure}[!h]
\scriptsize
\pgfplotsset{width=4.7cm,height=4cm,compat=1.8}
\begin{center}
\begin{tikzpicture}
    \begin{scope}[xshift=-1.0cm]
   \begin{axis}[
	 ymode=log,
	 grid=major,
     xmin = 1, xmax = 20,
     ymin = 0.0001, ymax = 0.02,
     axis y line* = left, 
     xlabel = {L-BFGS-B steps},
     title = {Monitoring \Gram}
     ]
     \addplot[blue,line width=1.1pt] coordinates {
( 1 , 0.011032095178961754 )
( 2 , 0.0037344691809266806 )
( 3 , 0.0019821743480861187 )
( 4 , 0.0012991288676857948 )
( 5 , 0.0009525723871774971 )
( 6 , 0.0007557183271273971 )
( 7 , 0.0006333743222057819 )
( 8 , 0.0005420579691417515 )
( 9 , 0.00048224107013083994 )
( 10 , 0.0004347219364717603 )
( 11 , 0.0003979738103225827 )
( 12 , 0.0003642793162725866 )
( 13 , 0.0003357716486789286 )
( 14 , 0.0003137352177873254 )
( 15 , 0.00029561552219092846 )
( 16 , 0.0002781266812235117 )
( 17 , 0.0002642460458446294 )
( 18 , 0.0002528093755245209 )
( 19 , 0.00024172077246475965 )
( 20 , 0.00023076956858858466 )
};
     \addplot[red,line width=1.1pt] coordinates {
( 1 , 0.014509769156575203 )
( 2 , 0.005623773206025362 )
( 3 , 0.0032881409861147404 )
( 4 , 0.0023650527000427246 )
( 5 , 0.0018689304124563932 )
( 6 , 0.0015394805232062936 )
( 7 , 0.0013480631168931723 )
( 8 , 0.0011786564718931913 )
( 9 , 0.0010472722351551056 )
( 10 , 0.0009402087889611721 )
( 11 , 0.000873691460583359 )
( 12 , 0.0008344611851498485 )
( 13 , 0.0007579259108752012 )
( 14 , 0.0007166120922192931 )
( 15 , 0.0006979264435358346 )
( 16 , 0.0006476993439719081 )
( 17 , 0.0006039823638275266 )
( 18 , 0.0005850586458109319 )
( 19 , 0.0005623167962767184 )
( 20 , 0.0005462238332256675 )
};
   \end{axis}
   \end{scope}
   \begin{scope}[xshift=3.15cm]
   \begin{axis}[
     ymode=log,
     grid=major,
     xmin = 1, xmax = 20,
     ymin = 0.001, ymax = 1.0,
     legend entries={{\hspace{-2mm}Minimize \Hist},
                     {Minimize \Gram},
                     },
	 legend style={
	   at={(0.2,0.95)},
	   anchor=north west
	 },
	 title = {Monitoring \Hist},
	 xlabel = {L-BFGS-B steps},
     ]
     \addplot[red,line width=1.1pt] coordinates {
( 1 , 0.021090922877192497 )
( 2 , 0.009767690673470497 )
( 3 , 0.0066136447712779045 )
( 4 , 0.00523045938462019 )
( 5 , 0.004543857648968697 )
( 6 , 0.0040597133338451385 )
( 7 , 0.003742966800928116 )
( 8 , 0.003472365438938141 )
( 9 , 0.0032730954699218273 )
( 10 , 0.0031499757897108793 )
( 11 , 0.0029697073623538017 )
( 12 , 0.0029117013327777386 )
( 13 , 0.0027969658840447664 )
( 14 , 0.002726455684751272 )
( 15 , 0.0026474285405129194 )
( 16 , 0.002613430144265294 )
( 17 , 0.002570465439930558 )
( 18 , 0.0025367322377860546 )
( 19 , 0.0024685340467840433 )
( 20 , 0.0023758569732308388 )
};
    \addplot[blue,line width=1.1pt] coordinates {
( 1 , 0.06159816309809685 )
( 2 , 0.04106029123067856 )
( 3 , 0.034440476447343826 )
( 4 , 0.03192000836133957 )
( 5 , 0.03075588122010231 )
( 6 , 0.030289528891444206 )
( 7 , 0.03013751469552517 )
( 8 , 0.029957886785268784 )
( 9 , 0.02970409020781517 )
( 10 , 0.029502511024475098 )
( 11 , 0.02960151433944702 )
( 12 , 0.029480453580617905 )
( 13 , 0.029726672917604446 )
( 14 , 0.029979977756738663 )
( 15 , 0.030313648283481598 )
( 16 , 0.030764136463403702 )
( 17 , 0.031052187085151672 )
( 18 , 0.031243622303009033 )
( 19 , 0.0315205380320549 )
( 20 , 0.031703103333711624 )
};
   \end{axis}
    \end{scope}
\end{tikzpicture}
\end{center}
\vspace{-6mm}
\caption{\label{fig:loss_convergence}
\textbf{Loss curves for the images of Figure~\ref{fig:texture_synthesis}.}
Optimizing for \Gram (blue) or \Hist (red).
We monitor the evolution of the values of \Gram (left) and \Hist (right) in either case.
\Hist encompasses \Gram: minimizing \Hist minimizes \Gram but the opposite is not true.
}
\vspace{-4mm}
\end{figure}

Because the loss \Hist captures the complete stationary statistics of deep features, it achieves the upperbound of what can be extracted as a stationary statistic from the layers of a given convolutional neural network. 
For instance, it encompasses \Gram~\cite{Gatys15}, $\mathcal{L}_{\tiny\text{Hist}}$~\cite{Risser17}, \Context~\cite{ContextualLoss} and \STROTSS~\cite{Kolkin_2019_CVPR}:
\begin{align}
\label{eq:swd_implication3}
\mathcal{L}_{\tiny\text{SW}}(I, \tilde I) = 0 &\implies \mathcal{L}_{\tiny\text{Gram}}(I, \tilde I) = 0, \\
\mathcal{L}_{\tiny\text{SW}}(I, \tilde I) = 0 &\implies \mathcal{L}_{\tiny\text{Hist}}(I, \tilde I) = 0, \\
\mathcal{L}_{\tiny\text{SW}}(I, \tilde I) = 0 &\implies \mathcal{L}_{\tiny\text{CX}}(I, \tilde I) = 0, \\
 \mathcal{L}_{\tiny\text{SW}}(I, \tilde I) = 0 &\implies \mathcal{L}_{\tiny\text{REMD}}(I, \tilde I) = 0
\end{align}
Figure~\ref{fig:loss_convergence} shows experimentally that optimizing for \Hist also optimizes for \Gram while the opposite is not true.

\paragraph{Implementation.}

Listing~\ref{lst:1} shows that it boils down to projecting the features on random directions (\ie unit vectors of 
dimension $N_l$), sort the projections and measure the $L^2$ distance on the sorted lists.
Because it computes a $L^2$ distance on a sorted list, this loss is differentiable everywhere and can be used for gradient retropropagation.
Furthermore, Tensorflow and Pytorch both provide an efficient GPU implementation of the \texttt{sort()} function.
To create images $n$ times larger than the example, we simply repeat each entry $n$ times in the latter's sorted list when evaluating the loss.

    \definecolor{lstshade}{gray}{0.95}
    \definecolor{lstframe}{gray}{0.80}
    \definecolor{lstcomment}{rgb}{0.0, 0.5, 0.0}
    \definecolor{lstkeyword}{rgb}{0.0, 0.0, 0.5}
    \definecolor{lstnumber}{rgb}{0.5, 0.0, 0.0}
    \definecolor{lstattrib}{rgb}{0,0.34,0}
    \lstset{
    	language=Python,
    	basicstyle = \scriptsize\ttfamily\raggedright,
    	commentstyle=\color{lstcomment}\itshape,
    	stringstyle=\color{lstattrib},
    	mathescape = true,
    	frame = lrtb,
    	backgroundcolor = \color{lstshade},
    	rulecolor = \color{lstframe},
    	tabsize = 3,
    	columns = fullflexible,
    	keepspaces,
    	belowskip = \smallskipamount,
    	framerule = .7pt,
    	breaklines = true,
    	showstringspaces = false,
    	keywordstyle = \color{lstkeyword},
    	numberstyle = \color{lstnumber},
    	captionpos = t,
    	texcl=false,
    	morekeywords={Conv2D,Lambda,K,reshape,tf,sort,shape,axis,mean},
    	caption={\textbf{Implementation of the Sliced Wasserstein loss.} The variable \texttt{Vs} is a matrix whose columns are normalized random directions in feature space. Random directions are redrawn for each batch.},
    	label={lst:1},
    	captionpos=b
    }
\begin{lstlisting}
# slicing
Vs = random_directions()
def Slicing(F):
	# project each pixel feature onto directions
	proj = dot(F, Vs)
	# flatten pixel indices to [M,N]
	H, W, N = proj.shape
	proj_flatten = reshape(proj,(H*W,N))
	# sort projections for each direction
	return sort(proj_flatten, axis=0)

# Sliced Wasserstein loss between two layers
def SlicedWassersteinLoss(F, F_):
	diff = Slicing(F) - Slicing(F_)
	return mean(square(diff))
\end{lstlisting}

\paragraph{Number of random directions.}
Iterating over random directions makes \Hist converge towards the target distribution regardless of the number of directions.
However, this number has an effect analogous to the batch size for stochastic gradient descent: it influences the noise in the gradient.
In Figure~\ref{fig:n_slices}, we compare convergence (monitored as \Hist with many directions) of optimisations that use different numbers of directions.
A low number is faster to compute and uses less memory but is slower to converge due to noisy gradients. 
A high number requires more computation and memory but generates less noisy gradients thus converges faster.
In practice, we use $N_l$ random directions, \ie as many as there are features in layer~$l$.
In this setting, we note an increased computational cost of $\approx 1.7-2.8 \times$ over \Gram.

\definecolor{darkred}{rgb}{0.86,0.00,0.00}
\begin{figure}[!h]
\scriptsize
\pgfplotsset{width=4.7cm,height=3.8cm,compat=1.8}
\begin{center}
\begin{tikzpicture}
   \begin{scope}[xshift=-1.0cm]
   \begin{axis}[
     ymode=log,
     grid=major,
     xmin = 1, xmax = 10,
	 legend style={
	   at={(1.02,1.05)},
	   anchor=north west
	 },
	 title style={yshift=2.0ex},
	 title = {Monitoring \Hist (2048 directions)},
	 xlabel = {L-BFGS-B steps},
     ]
     \addplot[pink,line width=1.1pt] coordinates { 
( 1 , 0.05619893 )
( 2 , 0.045946718 )
( 3 , 0.045165868 )
( 4 , 0.036930843 )
( 5 , 0.033454120000000004 )
( 6 , 0.038471908 )
( 7 , 0.033318732000000004 )
( 8 , 0.028302322999999997 )
( 9 , 0.02957479 )
( 10 , 0.026702095 )
};
     \addplot[magenta,line width=1.1pt] coordinates { 
( 1 , 0.021420543 )
( 2 , 0.016334763 )
( 3 , 0.015563484 )
( 4 , 0.01484666 )
( 5 , 0.013825342 )
( 6 , 0.012854817000000001 )
( 7 , 0.0120350946 )
( 8 , 0.011367876 )
( 9 , 0.011222542 )
( 10 , 0.0106011146 )
      };
    \addplot[darkred,line width=1.1pt] coordinates { 
( 1 , 0.009679717000000001 )
( 2 , 0.007881475 )
( 3 , 0.006946645999999999 )
( 4 , 0.006362748 )
( 5 , 0.0060248436 )
( 6 , 0.0057876587 )
( 7 , 0.0054667416 )
( 8 , 0.005367301 )
( 9 , 0.0052088410000000005 )
( 10 , 0.0051428007 )
     };
\addplot[purple,line width=1.1pt] coordinates { 
( 1 , 0.007636400000000001 )
( 2 , 0.005798159 )
( 3 , 0.005123031 )
( 4 , 0.0047521622 )
( 5 , 0.004525307 )
( 6 , 0.004344345 )
( 7 , 0.004232175 )
( 8 , 0.004135966 )
( 9 , 0.004054036 )
( 10 , 0.0039958682 )
     };
   \end{axis}
    \end{scope}
        \begin{scope}[xshift=3.15cm]
  \begin{axis}[
	 grid=major,
	 xtick={64, 128, 256, 512},
	 ytick={0, 100, 200, 300, 400},
     xmin = 32, xmax = 513,
     ymin = 0.0, ymax = 410.0,
     axis y line* = left, 
     xlabel = {Texture Size},
     title = {Time (seconds)},
     title style={yshift=2.0ex},
     legend entries={   {1 dir}\;\;\;,
                        {8 dirs\;\;\;},
                        {16 dirs\;\;\;},
                        {512 dirs}
                     },
	 legend columns=-1,
	 legend style={
	   inner xsep=0pt,inner ysep=0pt,nodes={inner sep=2pt,text depth=0.15em},
	   at={(-1.15,1.18)},
	   draw=none, 
	   anchor=north west
	 },
     ]
\addplot[pink,line width=1.1pt] coordinates { 
( 32 , 39.80227494239807 )
( 64 , 42.16823124885559 )
( 96 , 44.239017724990845 )
( 128 , 48.67606258392334 )
( 160 , 50.9332914352417 )
( 192 , 66.8473379611969 )
( 224 , 60.6148407459259 )
( 256 , 64.86556720733643 )
( 288 , 72.82508897781372 )
( 320 , 82.67773079872131 )
( 352 , 88.03312945365906 )
( 384 , 97.70858407020569 )
( 416 , 107.9210729598999 )
( 448 , 119.15454483032227 )
( 480 , 129.9303469657898 )
( 512 , 144.59954953193665 )
};
\addplot[magenta,line width=1.1pt] coordinates { 
( 32 , 36.29487895965576 )
( 64 , 40.833879470825195 )
( 96 , 42.1155903339386 )
( 128 , 44.378631830215454 )
( 160 , 50.24978709220886 )
( 192 , 52.435465574264526 )
( 224 , 58.92330312728882 )
( 256 , 65.84018850326538 )
( 288 , 74.34567260742188 )
( 320 , 80.76529169082642 )
( 352 , 91.99840140342712 )
( 384 , 103.34165668487549 )
( 416 , 116.51691102981567 )
( 448 , 129.11117100715637 )
( 480 , 142.9457242488861 )
( 512 , 157.43310832977295 )
};
\addplot[darkred,line width=1.1pt] coordinates { 
( 32 , 36.79343390464783 )
( 64 , 41.57185959815979 )
( 96 , 45.90934133529663 )
( 128 , 52.35758590698242 )
( 160 , 58.07971501350403 )
( 192 , 64.70691895484924 )
( 224 , 73.09675216674805 )
( 256 , 91.65638566017151 )
( 288 , 108.88317346572876 )
( 320 , 125.18880915641785 )
( 352 , 146.28125166893005 )
( 384 , 167.52087378501892 )
( 416 , 192.65440821647644 )
( 448 , 215.69335794448853 )
( 480 , 244.31195163726807 )
( 512 , 270.03340244293213 )
};
\addplot[purple,line width=1.1pt] coordinates { 
( 32 , 46.25670027732849 )
( 64 , 56.31552529335022 )
( 96 , 82.56546974182129 )
( 128 , 112.71744513511658 )
( 160 , 148.83864450454712 )
( 192 , 200.5076026916504 )
( 224 , 254.13658714294434 )
( 256 , 319.6889991760254 )
( 288 , 394.3867814540863 )
( 320 , 469.15955996513367 )
( 352 , 577.6697106361389 )
( 384 , 679.3151488304138 )
( 416 , 790.6183242797852 )
};
  \end{axis}
  \end{scope}

\end{tikzpicture}
\end{center}
\vspace{-6mm}
\caption{\textbf{Number of random directions.}
We compare convergence (left, texture size $256^2$) and runtime (right, various sizes, 10 steps) of optimizing \Hist with a varying number of directions.
Setting: SciPy's L-BFGS-B ($maxfun=64, pgtol=0.0, factr=0.0$) in Python and Tensorflow 2.3 on Intel Core i5 and NVidia Titan Xp.
\label{fig:n_slices}
}
\vspace{-6mm}
\end{figure}
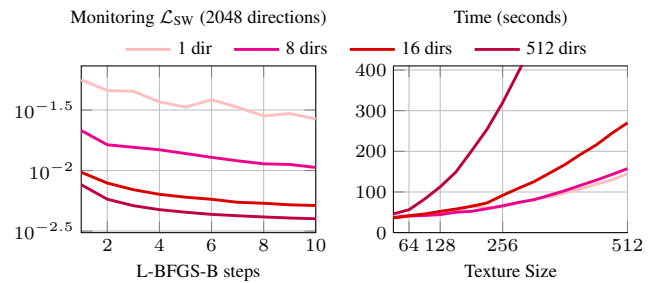

\paragraph{Texture synthesis by optimization.}

\begin{figure*}[h]
\vspace{-5mm}
\begin{center}
\begin{tabular}{@{} c @{\hspace{1mm}} c @{\hspace{1mm}} c @{\hspace{4mm}} c @{\hspace{1mm}} c @{\hspace{1mm}} c @{}}
& 
\begin{tabular}{@{} c @{}}
\scriptsize\textbf{optim} {\small \Gram} \vspace{-2mm}\\
\scriptsize \scalebox{0.75}{$512\times512$}\\
\end{tabular}
& 
\begin{tabular}{@{} c @{}}
\scriptsize\textbf{optim} {\small \Hist} \vspace{-2mm}\\
\scriptsize \scalebox{0.75}{$512\times512$}\\
\end{tabular} 
&
& 
\begin{tabular}{@{} c @{}}
\scriptsize\textbf{optim} {\small \Gram} \vspace{-2mm}\\
\scriptsize \scalebox{0.75}{$512\times512$}\\
\end{tabular}
& 
\begin{tabular}{@{} c @{}}
\scriptsize\textbf{optim} {\small \Hist} \vspace{-2mm}\\
\scriptsize \scalebox{0.75}{$512\times512$}\\
\end{tabular} 
\\
\raisebox{17mm}{
\begin{tabular}{@{} c @{}}
\scriptsize\textbf{input} \vspace{-2mm} \\
\scriptsize \scalebox{0.75}{$128\times128$}\\
\fbox{\includegraphics[width=0.045\linewidth]{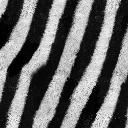}}
\end{tabular}
}
&
\fbox{\includegraphics[width=0.18\linewidth]{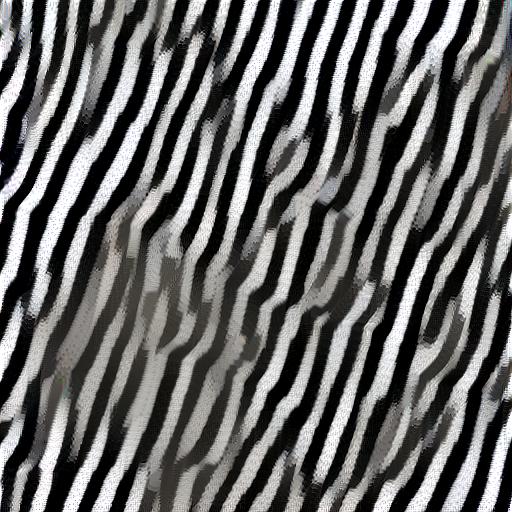}}&
\fbox{\includegraphics[width=0.18\linewidth]{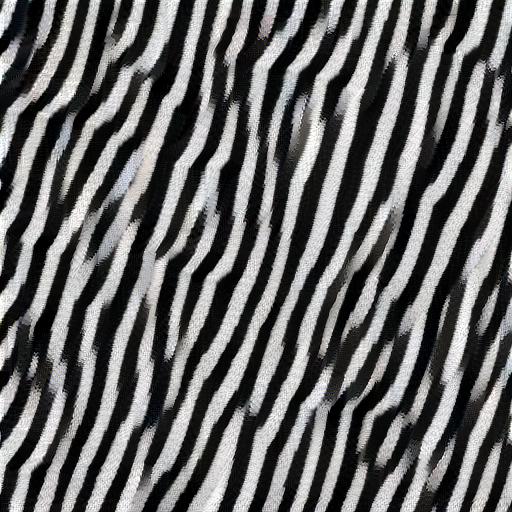}}
&
\raisebox{17mm}{
\begin{tabular}{@{} c @{}}
\scriptsize\textbf{input} \vspace{-2mm} \\
\scriptsize \scalebox{0.75}{$256\times256$}\\
\fbox{\includegraphics[width=0.09\linewidth]{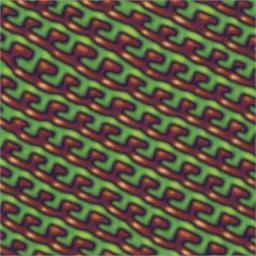}}
\end{tabular}
}&
\fbox{\includegraphics[width=0.18\linewidth]{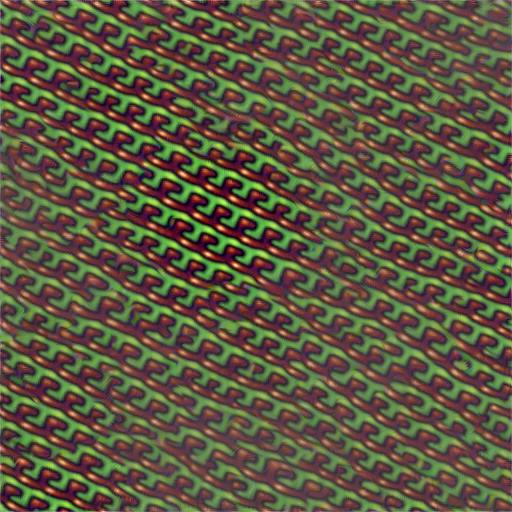}}&
\fbox{\includegraphics[width=0.18\linewidth]{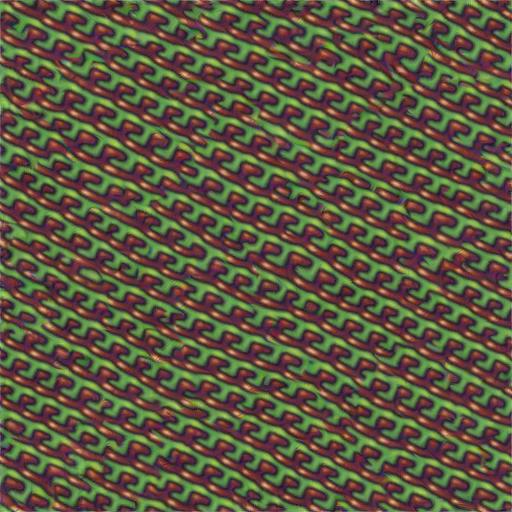}}
\\
\raisebox{17mm}{
\begin{tabular}{@{} c @{}}
\scriptsize\textbf{input} \vspace{-2mm} \\
\scriptsize \scalebox{0.75}{$128\times128$}\\
\fbox{\includegraphics[width=0.045\linewidth]{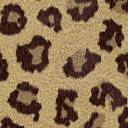}}
\end{tabular}
}
&
\fbox{\includegraphics[width=0.18\linewidth]{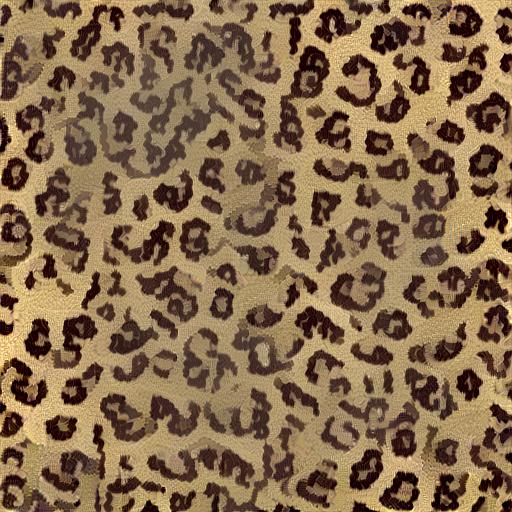}}&
\fbox{\includegraphics[width=0.18\linewidth]{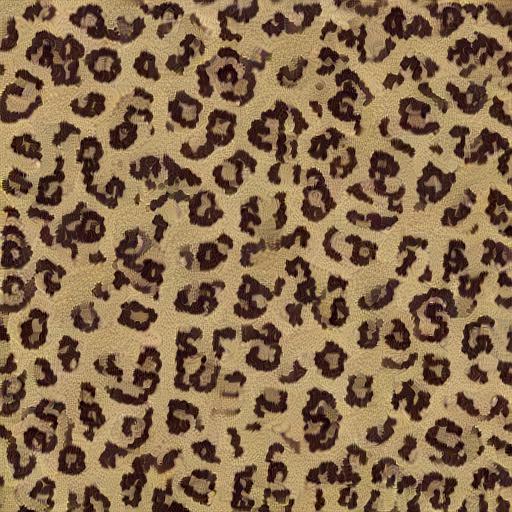}}
&
\raisebox{17mm}{
\begin{tabular}{@{} c @{}}
\scriptsize\textbf{input} \vspace{-2mm} \\
\scriptsize \scalebox{0.75}{$256\times256$}\\
\fbox{\includegraphics[width=0.09\linewidth]{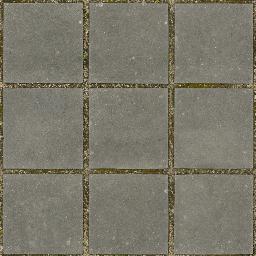}}
\end{tabular}
}
&
\fbox{\includegraphics[width=0.18\linewidth]{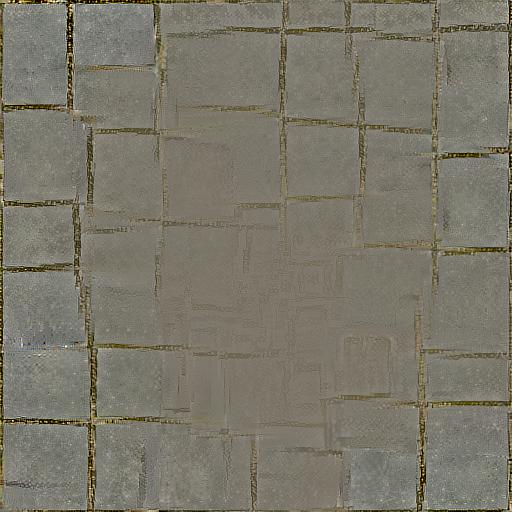}}&
\fbox{\includegraphics[width=0.18\linewidth]{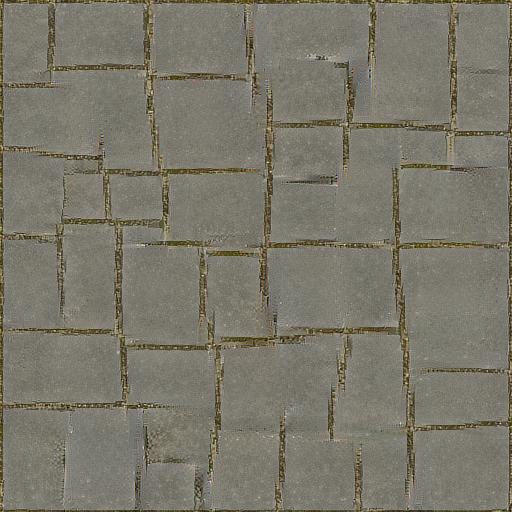}}
\\
\raisebox{17mm}{
\begin{tabular}{@{} c @{}}
\scriptsize\textbf{input} \vspace{-2mm} \\
\scriptsize \scalebox{0.75}{$128\times128$}\\
\fbox{\includegraphics[width=0.045\linewidth]{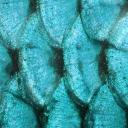}}
\end{tabular}
}
&
\fbox{\includegraphics[width=0.18\linewidth]{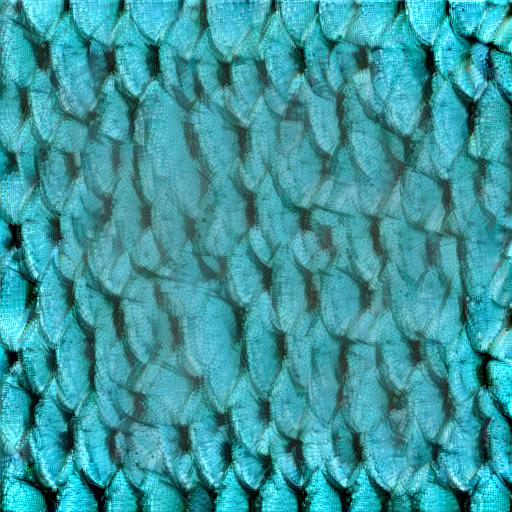}}&
\fbox{\includegraphics[width=0.18\linewidth]{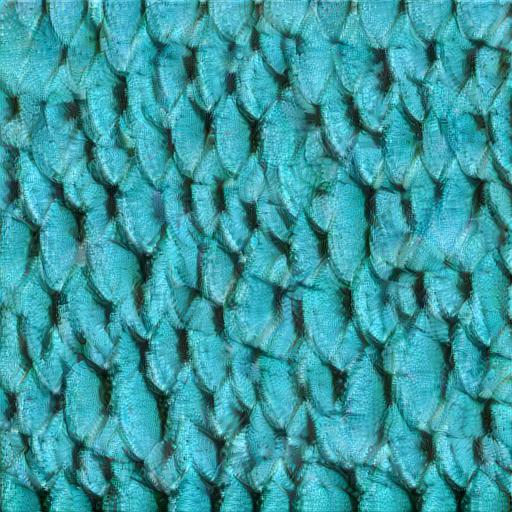}}&
%
\raisebox{17mm}{
\begin{tabular}{@{} c @{}}
\scriptsize\textbf{input} \vspace{-2mm} \\
\scriptsize \scalebox{0.75}{$256\times256$}\\
\fbox{\includegraphics[width=0.09\linewidth]{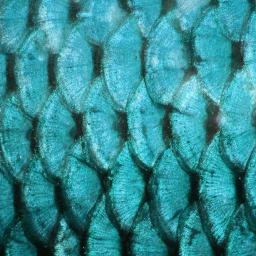}}
\end{tabular}
}
&
\fbox{\includegraphics[width=0.18\linewidth]{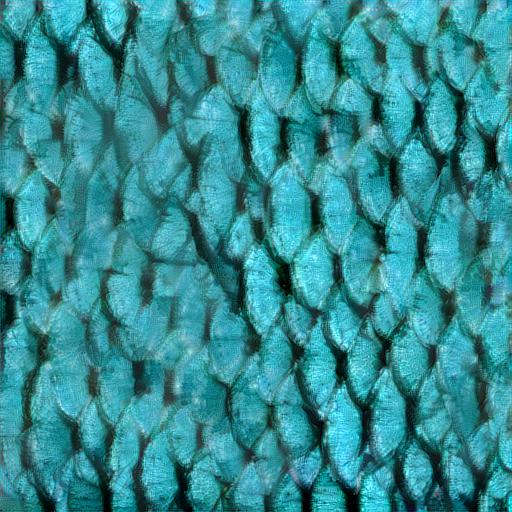}}&
\fbox{\includegraphics[width=0.18\linewidth]{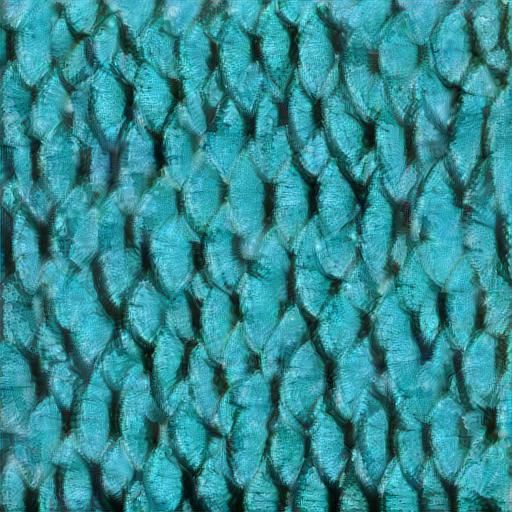}}
\end{tabular}
\end{center}
\vspace{-6mm}
\caption{\label{fig:texture_synthesis} \textbf{Texture synthesis by optimization.} 
We optimize a Gaussian white noise with \Gram and \Hist using L-BGFS. 
In the last row, the inputs are two crops of different sizes extracted from the same texture.
}
\vspace{-2mm}
\end{figure*}

\begin{figure*}[!h]
\begin{center}
\begin{tabular}{@{\hspace{-0.3mm}} c @{\hspace{1.5mm}} c @{\hspace{1mm}} c @{\hspace{1mm}} c @{\hspace{1.5mm}} c @{\hspace{1mm}} c @{\hspace{1mm}} c @{}}
\scriptsize\textbf{Content} &
\scriptsize\textbf{Style} &
\scriptsize\textbf{optim \Gram} &
\scriptsize\textbf{optim \Hist} &
\scriptsize\textbf{Style} &
\scriptsize\textbf{optim \Gram} &
\scriptsize\textbf{optim \Hist} 
\vspace{-2mm} \\
\scriptsize \scalebox{0.75}{$512\times512$} &
\scriptsize \scalebox{0.75}{$128\times128$} &
\scriptsize \scalebox{0.75}{$512\times512$} &
\scriptsize \scalebox{0.75}{$512\times512$} &
\scriptsize \scalebox{0.75}{$256\times256$} &
\scriptsize \scalebox{0.75}{$512\times512$} &
\scriptsize \scalebox{0.75}{$512\times512$} 
\\
\fbox{\includegraphics[width=0.16\linewidth]{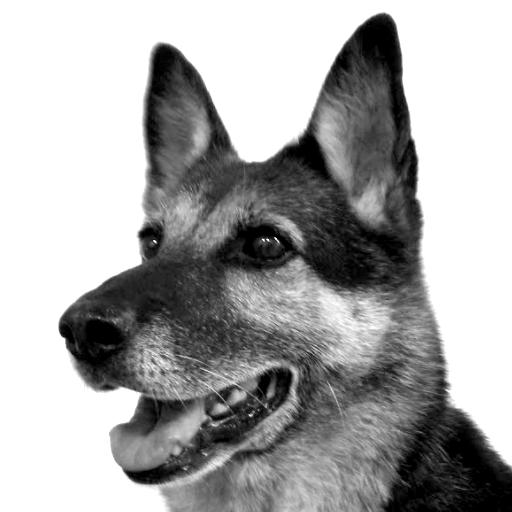}} &
\raisebox{11mm}{
\fbox{\includegraphics[width=0.04\linewidth]{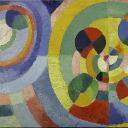}}
}&
\fbox{\includegraphics[width=0.16\linewidth]{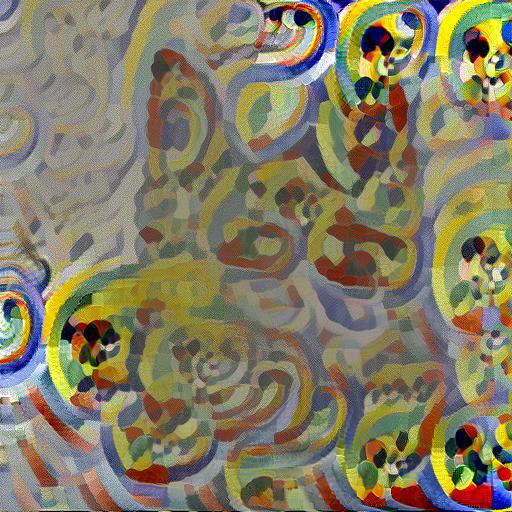}} &
\fbox{\includegraphics[width=0.16\linewidth]{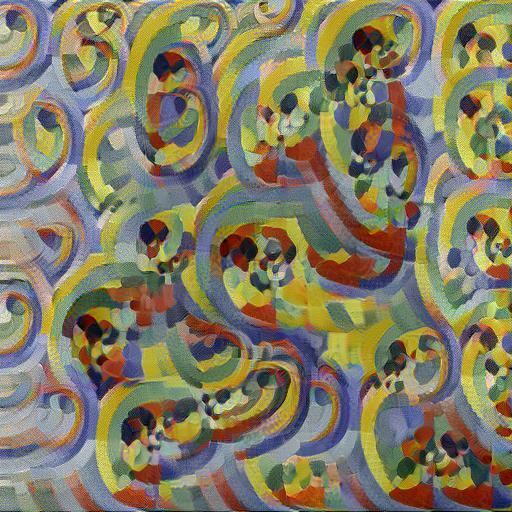}} &
\raisebox{7mm}{
\fbox{\includegraphics[width=0.08\linewidth]{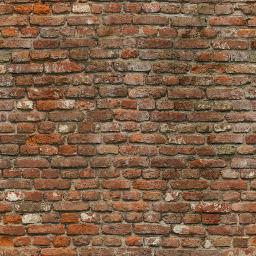}}
}
&
\fbox{\includegraphics[width=0.16\linewidth]{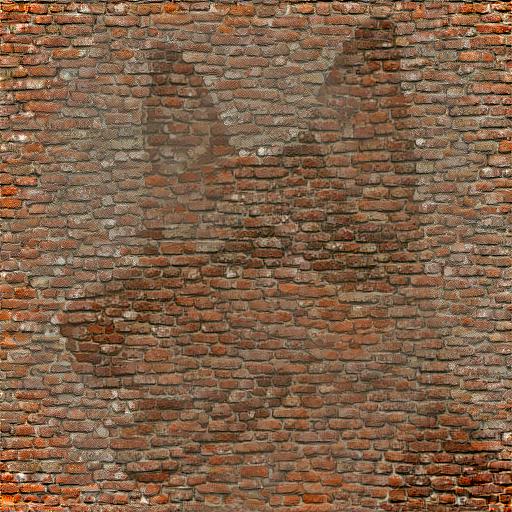}} &
\fbox{\includegraphics[width=0.16\linewidth]{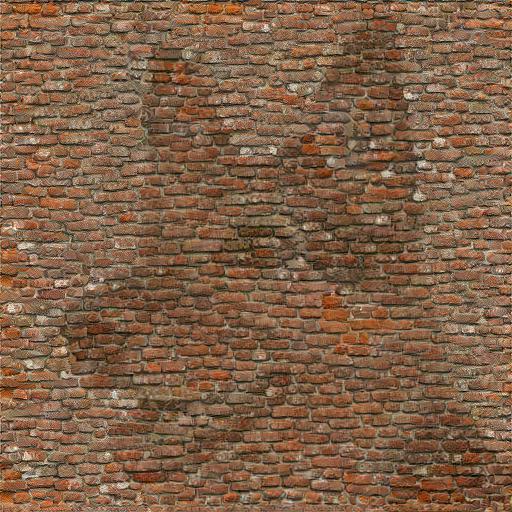}}
\end{tabular}
\end{center}
\vspace{-6mm}
\caption{\label{fig:style_transfer} \textbf{Style transfer by optimization.} 
We optimize the content image with \Gram and \Hist using L-BGFS (we do not add a content loss with a tuneable parameter such as~\cite{Gatys16}). }
\vspace{-5mm}
\end{figure*}

In Figure~\ref{fig:texture_synthesis}, we compare \Hist to \Gram in the scope of iterative texture optimization with an L-BFGS-B optimizer.
This setting is the right unit test to validate the textural loss that drives the optimization, avoiding issues related to training neural networks or meta-parameter tuning.
We observe that \Gram produces artifacts such as oscillating contrasts and is inconsistent \wrt different input sizes (last row), making it less predictable.
These limitations of \Gram are documented in previous works~\cite{Risser17,DeepCorrelation}.
In contrast, \Hist generates uniformly sharp textures consistent \wrt different input sizes.
Figure~\ref{fig:style_transfer} confirms these observations with style transfer. 

\vspace{-1mm}

\paragraph{Training generative neural networks.}

\begin{figure*}[h]
\centering
\begin{tabular}{@{} c @{\hspace{0.5mm}} c @{\hspace{1mm}} c @{\hspace{0.5mm}} c @{\hspace{1mm}} c @{\hspace{0.5mm}} c @{}}
{\scriptsize\textbf{example}} & {\scriptsize\textbf{generated}} &
{\scriptsize\textbf{example}} & {\scriptsize\textbf{generated}} &
{\scriptsize\textbf{example}} & {\scriptsize\textbf{generated}} \\
\fbox{\includegraphics[height=1.6cm]{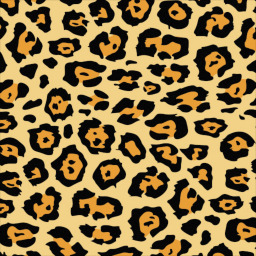}} &
\fbox{\includegraphics[height=1.6cm, trim=1415 0 0 0, clip]{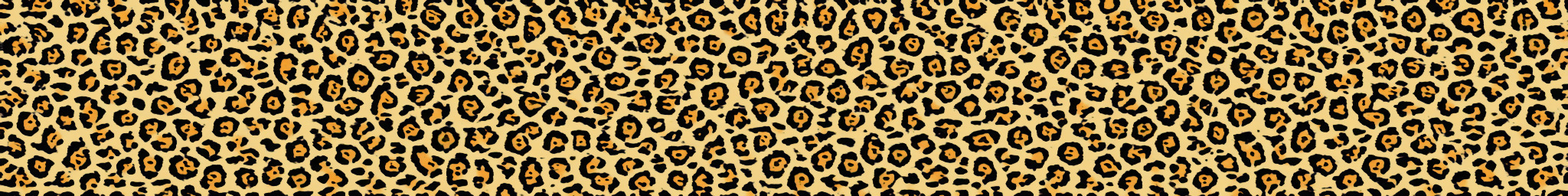}} &
\fbox{\includegraphics[height=1.6cm]{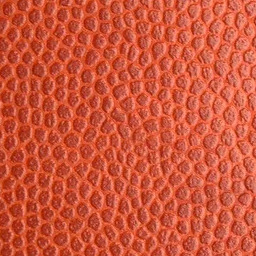}} &
\fbox{\includegraphics[height=1.6cm, trim=1415 0 0 0, clip]{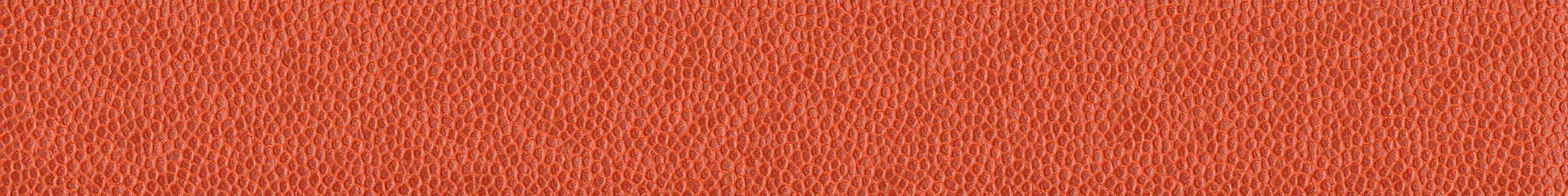}} &
\fbox{\includegraphics[height=1.6cm]{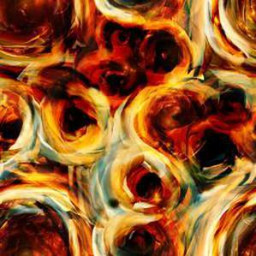}} &
\fbox{\includegraphics[height=1.6cm, trim=1415 0 0 0, clip]{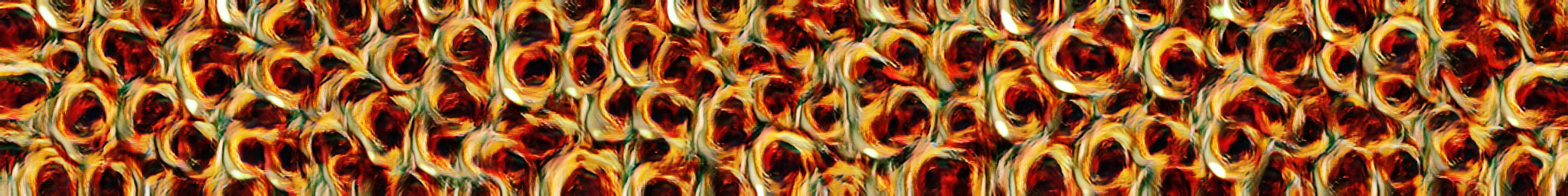}} 
\end{tabular}
\vspace{-3.5mm}
\caption{\label{fig:mono_texture} \textbf{Training a mono-texture generator.}
We use \Hist to train a mono-texture generative architecture~\cite{UlyanovV1}.
}
\vspace{-5mm}
\end{figure*}

\begin{figure*}[h]
\centering
\begin{tabular}{@{} c @{\hspace{1mm}} c @{\hspace{1mm}} c @{\hspace{2mm}}  c @{\hspace{1mm}}  c @{\hspace{1mm}}  c @{\hspace{1mm}} @{}}
{\scriptsize\textbf{example 1}} & {\scriptsize\textbf{generated}} & {\scriptsize\textbf{example 2}} & {\scriptsize\textbf{example 3}} & {\scriptsize\textbf{generated}} & {\scriptsize\textbf{example 4}} \\
\fbox{\includegraphics[height=1.6cm]{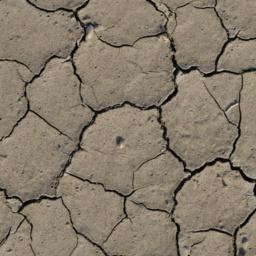}} &
\fbox{\includegraphics[height=1.6cm]{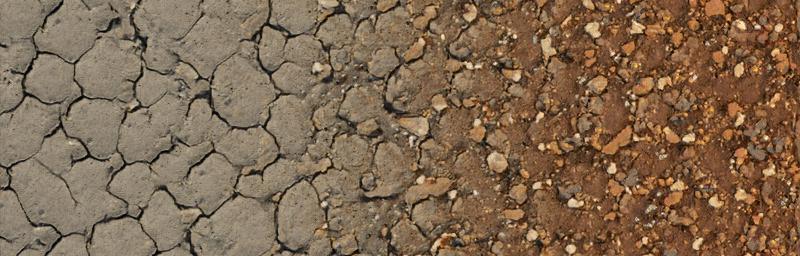}} &
\fbox{\includegraphics[height=1.6cm]{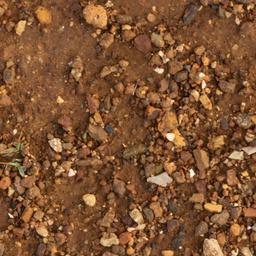}} &
\fbox{\includegraphics[height=1.6cm]{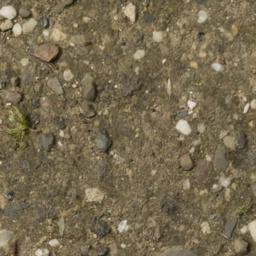}} &
\fbox{\includegraphics[height=1.6cm]{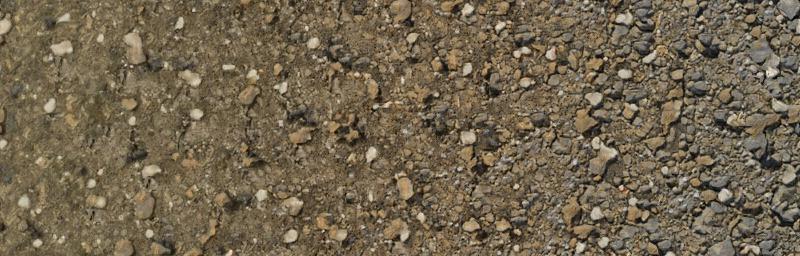}} &
\fbox{\includegraphics[height=1.6cm]{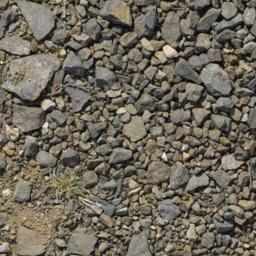}}
\end{tabular}
\vspace{-4mm}
\caption{\label{fig:multi_texture} \textbf{Training a multi-texture generator.}
We use \Hist to train a multi-texture generative architecture that allows for interpolation~\cite{Li2017}. 
We trained the same architecture for 32 textures that includes these 4 examples.
}
\vspace{-4mm}
\end{figure*}

While direct texture optimization is a good unit test for the loss, we also validate that we can successfully use \Hist for training.
In Figure~\ref{fig:mono_texture} and~\ref{fig:multi_texture}
we use \Hist for sole loss function to train a mono-texture~\cite{UlyanovV1} and a multi-texture~\cite{Li2017} generative architecture, respectively. 
They are capable of producing arbitrarily-large texture at inference time, with variation (no verbatim copying of the exemplar) and interpolation.
This experiment validates that there are \textit{a priori} no obstacles to using \Hist as a drop-in replacement for \Gram for training.

\clearpage

\setlength{\fboxsep}{0pt}\setlength{\fboxrule}{0.8pt}

\setlength{\fboxsep}{0pt}\setlength{\fboxrule}{0.8pt}

\clearpage

\clearpage

\section{Spatial Constraints Via User-Defined Tags}
\label{sec:spatial}

The Sliced Wasserstein loss \Hist presented in Section~\ref{sec:contrib} captures all the stationary, \ie position-agnostic, statistics.
Like the Gram loss, it means that one has no spatial control over the synthesized textures that are optimized with this loss.
For some applications where spatial constraints are required, previous works usually use additional losses whose relative weighting has to be fine-tuned. 
In this section, we propose a simple way to incorporate spatial constraints in the Sliced Wasserstein loss \Hist without any modification and without adding further losses.

\setlength{\fboxsep}{0pt}\setlength{\fboxrule}{0.8pt}
\begin{figure}[!h]
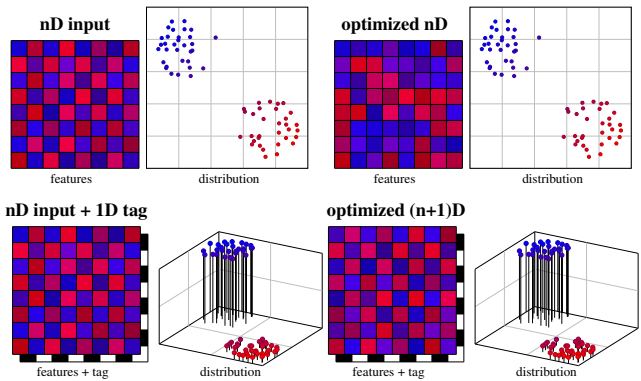

\begin{center}
\hspace{-5mm}
\begin{tikzpicture}
    \begin{scope}
        \draw (0.85, 1.4) node[anchor=center] {\scriptsize \textbf{nD input}};
        \begin{scope}[shift={(0.0,-0.5)}, scale=1.7]
            \input{figure4_image1.tex}
        \end{scope}
    
        \begin{scope}[shift={(1.8,-0.5)}, scale=0.3]
    	    \begin{axis}[width=0.5\textwidth,height=0.5\textwidth,grid=major,xmin=0.0,xmax=1.0,ymin=0.0,ymax=1.0,ticks=none]
	\addplot[only marks, scatter,colormap name=mine,point meta=y-x,domain=0:1] 
		coordinates {
( 0.7375242851479528 , 0.07149719324412154 )
( 0.1352616126167896 , 0.8741453717776989 )
( 0.6659961465969623 , 0.16905674183969271 )
( 0.25811608155106386 , 0.6634410081431301 )
( 0.8163931968005627 , 0.14827914918912766 )
( 0.235522015159636 , 0.862679580008819 )
( 0.9358410694203565 , 0.23491796054853353 )
( 0.09431609762856191 , 0.808318117172903 )
( 0.1154706312820831 , 0.7728239727237151 )
( 0.8139804373163412 , 0.4008613797149576 )
( 0.157427062376505 , 0.9106254691891366 )
( 0.8087453710717747 , 0.0873358066465128 )
( 0.22286168569889056 , 0.6672178044029601 )
( 0.7229799035725907 , 0.11560227135707851 )
( 0.35162333787807665 , 0.6199816890366898 )
( 0.6992643553827685 , 0.19942557445733575 )
( 0.8817663778657872 , 0.2412614969652668 )
( 0.12451618465828287 , 0.7369979876185243 )
( 0.6097974800796597 , 0.35128155099802394 )
( 0.10479225209789506 , 0.6579663171400711 )
( 0.7009834334711795 , 0.3720120811638443 )
( 0.27739610562333833 , 0.8599495053784584 )
( 0.8201340945592945 , 0.22151647062355037 )
( 0.131654833648978 , 0.816398412146849 )
( 0.1245513907080335 , 0.630094832814329 )
( 0.9306947276438879 , 0.26962780887737386 )
( 0.1972961888949818 , 0.7946443747181786 )
( 0.8941919835488764 , 0.3450755917066442 )
( 0.08979436903883525 , 0.7668576885426852 )
( 0.8482706859776603 , 0.404644785626021 )
( 0.4284911662411657 , 0.8111718300538617 )
( 0.9076415266589243 , 0.15430439746777957 )
( 0.6610237230913176 , 0.33686074688022727 )
( 0.2701167616942569 , 0.8142574315792738 )
( 0.6539149012676253 , 0.14837209219747058 )
( 0.2841292635125761 , 0.7305270744563124 )
( 0.9277902148119016 , 0.1938935051088681 )
( 0.27391219388005256 , 0.5725148987449746 )
( 0.8655132690749044 , 0.2713106951700408 )
( 0.0734335217243964 , 0.7396019784732524 )
( 0.15439783421579517 , 0.58650271025823 )
( 0.6512898381036001 , 0.2000216613180242 )
( 0.1995156704890989 , 0.8120878635117923 )
( 0.821622311898793 , 0.2790205918520516 )
( 0.19624086371465507 , 0.5807610872932932 )
( 0.8636534200400908 , 0.3083595677085642 )
( 0.2679208190322526 , 0.6244303950353428 )
( 0.6038967380125001 , 0.36810405277789554 )
( 0.8033544609775752 , 0.36867725590955347 )
( 0.2740573751979177 , 0.7243578756697717 )
( 0.6948557533715101 , 0.18925959140395698 )
( 0.17285575256546654 , 0.7679928808749046 )
( 0.8647811959996955 , 0.18825570536856379 )
( 0.23518869693404476 , 0.9128898340829545 )
( 0.765926429611197 , 0.4133710202209234 )
( 0.12201069060200645 , 0.6219079342588435 )
( 0.19021263915389325 , 0.7348070328372469 )
( 0.5813158904524919 , 0.19543314629004738 )
( 0.2832485755015384 , 0.7709581697935483 )
( 0.6970072406622857 , 0.40678203883304376 )
( 0.2340980139455316 , 0.7627891475368583 )
( 0.7254402658099074 , 0.3060292568519621 )
( 0.16488620203439142 , 0.8593651794811017 )
( 0.7027023418778426 , 0.14055717654434494 )
       };
	\end{axis}
    	\end{scope}

        \draw[anchor=north] (0.8,-0.45) node {\tiny features};
        \draw[anchor=north] (2.9,-0.45) node {\tiny distribution};
	\end{scope}
\hspace{3mm}
    \begin{scope}[shift={(4.,0.0)}]
        \draw (0.8, 1.4) node[anchor=center] {\scriptsize \textbf{optimized nD}};
        \begin{scope}[shift={(0.0,-0.5)}, scale=1.7]
            \input{figure1_image.tex}
        \end{scope}
    
        \begin{scope}[shift={(1.8,-0.5)}, scale=0.3]
    	    \begin{axis}[width=0.5\textwidth,height=0.5\textwidth,grid=major,xmin=0.0,xmax=1.0,ymin=0.0,ymax=1.0,ticks=none]
	\addplot[only marks, scatter,colormap name=mine,point meta=y-x,domain=0:1] 
		coordinates {
( 0.7375242851479528 , 0.07149719324412154 )
( 0.1352616126167896 , 0.8741453717776989 )
( 0.6659961465969623 , 0.16905674183969271 )
( 0.25811608155106386 , 0.6634410081431301 )
( 0.8163931968005627 , 0.14827914918912766 )
( 0.235522015159636 , 0.862679580008819 )
( 0.9358410694203565 , 0.23491796054853353 )
( 0.09431609762856191 , 0.808318117172903 )
( 0.1154706312820831 , 0.7728239727237151 )
( 0.8139804373163412 , 0.4008613797149576 )
( 0.157427062376505 , 0.9106254691891366 )
( 0.8087453710717747 , 0.0873358066465128 )
( 0.22286168569889056 , 0.6672178044029601 )
( 0.7229799035725907 , 0.11560227135707851 )
( 0.35162333787807665 , 0.6199816890366898 )
( 0.6992643553827685 , 0.19942557445733575 )
( 0.8817663778657872 , 0.2412614969652668 )
( 0.12451618465828287 , 0.7369979876185243 )
( 0.6097974800796597 , 0.35128155099802394 )
( 0.10479225209789506 , 0.6579663171400711 )
( 0.7009834334711795 , 0.3720120811638443 )
( 0.27739610562333833 , 0.8599495053784584 )
( 0.8201340945592945 , 0.22151647062355037 )
( 0.131654833648978 , 0.816398412146849 )
( 0.1245513907080335 , 0.630094832814329 )
( 0.9306947276438879 , 0.26962780887737386 )
( 0.1972961888949818 , 0.7946443747181786 )
( 0.8941919835488764 , 0.3450755917066442 )
( 0.08979436903883525 , 0.7668576885426852 )
( 0.8482706859776603 , 0.404644785626021 )
( 0.4284911662411657 , 0.8111718300538617 )
( 0.9076415266589243 , 0.15430439746777957 )
( 0.6610237230913176 , 0.33686074688022727 )
( 0.2701167616942569 , 0.8142574315792738 )
( 0.6539149012676253 , 0.14837209219747058 )
( 0.2841292635125761 , 0.7305270744563124 )
( 0.9277902148119016 , 0.1938935051088681 )
( 0.27391219388005256 , 0.5725148987449746 )
( 0.8655132690749044 , 0.2713106951700408 )
( 0.0734335217243964 , 0.7396019784732524 )
( 0.15439783421579517 , 0.58650271025823 )
( 0.6512898381036001 , 0.2000216613180242 )
( 0.1995156704890989 , 0.8120878635117923 )
( 0.821622311898793 , 0.2790205918520516 )
( 0.19624086371465507 , 0.5807610872932932 )
( 0.8636534200400908 , 0.3083595677085642 )
( 0.2679208190322526 , 0.6244303950353428 )
( 0.6038967380125001 , 0.36810405277789554 )
( 0.8033544609775752 , 0.36867725590955347 )
( 0.2740573751979177 , 0.7243578756697717 )
( 0.6948557533715101 , 0.18925959140395698 )
( 0.17285575256546654 , 0.7679928808749046 )
( 0.8647811959996955 , 0.18825570536856379 )
( 0.23518869693404476 , 0.9128898340829545 )
( 0.765926429611197 , 0.4133710202209234 )
( 0.12201069060200645 , 0.6219079342588435 )
( 0.19021263915389325 , 0.7348070328372469 )
( 0.5813158904524919 , 0.19543314629004738 )
( 0.2832485755015384 , 0.7709581697935483 )
( 0.6970072406622857 , 0.40678203883304376 )
( 0.2340980139455316 , 0.7627891475368583 )
( 0.7254402658099074 , 0.3060292568519621 )
( 0.16488620203439142 , 0.8593651794811017 )
( 0.7027023418778426 , 0.14055717654434494 )
       };
	\end{axis}
    	\end{scope}
    	
        \draw[anchor=north] (0.8,-0.45) node {\tiny features};
        \draw[anchor=north] (2.9,-0.45) node {\tiny distribution};
	\end{scope}
	\end{tikzpicture}
	\\
	\hspace{-8mm}
	\begin{tikzpicture}
    \begin{scope}[shift={(6.5,0.0)}]
    \begin{scope}[shift={(0.0,0.0)}]
        \draw (0.85, 1.5) node[anchor=center] {\scriptsize \textbf{nD input + 1D tag}};
        
        \begin{scope}[shift={(0.1,-0.5)}, scale=1.7]
            \input{figure4_image4b.tex}
        \end{scope}
        \begin{scope}[shift={(0.0,-0.4)}, scale=1.7]
            \input{figure4_image1.tex}
        \end{scope}
    
        \begin{scope}[shift={(1.95,-0.55)}, scale=0.4]
    	    \begin{axis}[grid=major,
             view={210}{30},
             ticks=none,
            ]
	\addplot3[only marks, ycomb, scatter,colormap name=mine,point meta=x-y,domain=0:1] 
		coordinates {
( 0.7375242851479528 , 0.07149719324412154 , 0.8 )
( 0.6352616126167896 , 0.3741453717776989 , 0.8 )
( 0.6659961465969623 , 0.16905674183969271 , 0.8 )
( 0.7581160815510639 , 0.1634410081431301 , 0.8 )
( 0.8163931968005627 , 0.14827914918912766 , 0.8 )
( 0.235522015159636 , 0.862679580008819 , 0.2 )
( 0.4358410694203565 , 0.7349179605485335 , 0.2 )
( 0.09431609762856191 , 0.808318117172903 , 0.2 )
( 0.1154706312820831 , 0.7728239727237151 , 0.2 )
( 0.3139804373163412 , 0.9008613797149576 , 0.2 )
( 0.157427062376505 , 0.9106254691891366 , 0.2 )
( 0.8087453710717747 , 0.0873358066465128 , 0.8 )
( 0.22286168569889056 , 0.6672178044029601 , 0.2 )
( 0.22297990357259065 , 0.6156022713570786 , 0.2 )
( 0.8516233378780766 , 0.11998168903668985 , 0.8 )
( 0.19926435538276854 , 0.6994255744573358 , 0.2 )
( 0.38176637786578715 , 0.7412614969652668 , 0.2 )
( 0.12451618465828287 , 0.7369979876185243 , 0.2 )
( 0.10979748007965975 , 0.8512815509980239 , 0.2 )
( 0.10479225209789506 , 0.6579663171400711 , 0.2 )
( 0.7009834334711795 , 0.3720120811638443 , 0.8 )
( 0.7773961056233383 , 0.3599495053784583 , 0.8 )
( 0.8201340945592945 , 0.22151647062355037 , 0.8 )
( 0.131654833648978 , 0.816398412146849 , 0.2 )
( 0.6245513907080336 , 0.13009483281432893 , 0.8 )
( 0.430694727643888 , 0.7696278088773739 , 0.2 )
( 0.1972961888949818 , 0.7946443747181786 , 0.2 )
( 0.8941919835488764 , 0.3450755917066442 , 0.8 )
( 0.5897943690388352 , 0.26685768854268516 , 0.8 )
( 0.8482706859776603 , 0.404644785626021 , 0.8 )
( 0.4284911662411657 , 0.8111718300538617 , 0.2 )
( 0.4076415266589243 , 0.6543043974677796 , 0.2 )
( 0.16102372309131766 , 0.8368607468802273 , 0.2 )
( 0.2701167616942569 , 0.8142574315792738 , 0.2 )
( 0.1539149012676253 , 0.6483720921974706 , 0.2 )
( 0.2841292635125761 , 0.7305270744563124 , 0.2 )
( 0.9277902148119016 , 0.1938935051088681 , 0.8 )
( 0.27391219388005256 , 0.5725148987449746 , 0.2 )
( 0.36551326907490445 , 0.7713106951700408 , 0.2 )
( 0.0734335217243964 , 0.7396019784732524 , 0.2 )
( 0.6543978342157951 , 0.08650271025822998 , 0.8 )
( 0.6512898381036001 , 0.2000216613180242 , 0.8 )
( 0.1995156704890989 , 0.8120878635117923 , 0.2 )
( 0.821622311898793 , 0.2790205918520516 , 0.8 )
( 0.6962408637146551 , 0.08076108729329323 , 0.8 )
( 0.3636534200400908 , 0.8083595677085642 , 0.2 )
( 0.2679208190322526 , 0.6244303950353428 , 0.2 )
( 0.10389673801250013 , 0.8681040527778956 , 0.2 )
( 0.8033544609775752 , 0.36867725590955347 , 0.8 )
( 0.2740573751979177 , 0.7243578756697717 , 0.2 )
( 0.6948557533715101 , 0.18925959140395698 , 0.8 )
( 0.6728557525654665 , 0.2679928808749045 , 0.8 )
( 0.8647811959996955 , 0.18825570536856379 , 0.8 )
( 0.23518869693404476 , 0.9128898340829545 , 0.2 )
( 0.265926429611197 , 0.9133710202209234 , 0.2 )
( 0.6220106906020064 , 0.1219079342588435 , 0.8 )
( 0.6902126391538932 , 0.23480703283724697 , 0.8 )
( 0.08131589045249188 , 0.6954331462900474 , 0.2 )
( 0.7832485755015384 , 0.2709581697935482 , 0.8 )
( 0.19700724066228567 , 0.9067820388330438 , 0.2 )
( 0.7340980139455315 , 0.2627891475368584 , 0.8 )
( 0.22544026580990742 , 0.806029256851962 , 0.2 )
( 0.6648862020343914 , 0.3593651794811017 , 0.8 )
( 0.20270234187784264 , 0.640557176544345 , 0.2 )
       };
	\end{axis}
    	\end{scope}
        \draw[anchor=north] (0.8,-0.45) node {\tiny features + tag};
        \draw[anchor=north] (2.9,-0.45) node {\tiny distribution};
	\end{scope}
\hspace{5mm}
    \begin{scope}[shift={(3.70,0.0)}]
        \draw (0.9, 1.5) node[anchor=center] {\scriptsize \textbf{optimized (n+1)D}};
        
        \begin{scope}[shift={(0.1,-0.5)}, scale=1.7]
            \input{figure4_image4b.tex}
        \end{scope}
        \begin{scope}[shift={(0.0,-0.4)}, scale=1.7]
            \input{figure4_image4.tex}
        \end{scope}
    
        \begin{scope}[shift={(1.95,-0.55)}, scale=0.4]
    	    \begin{axis}[grid=major,
             view={210}{30},
             ticks=none,
            ]
	\addplot3[only marks, ycomb, scatter,colormap name=mine,point meta=x-y,domain=0:1] 
		coordinates {
( 0.7375242851479528 , 0.07149719324412154 , 0.8 )
( 0.6352616126167896 , 0.3741453717776989 , 0.8 )
( 0.6659961465969623 , 0.16905674183969271 , 0.8 )
( 0.7581160815510639 , 0.1634410081431301 , 0.8 )
( 0.8163931968005627 , 0.14827914918912766 , 0.8 )
( 0.235522015159636 , 0.862679580008819 , 0.2 )
( 0.4358410694203565 , 0.7349179605485335 , 0.2 )
( 0.09431609762856191 , 0.808318117172903 , 0.2 )
( 0.1154706312820831 , 0.7728239727237151 , 0.2 )
( 0.3139804373163412 , 0.9008613797149576 , 0.2 )
( 0.157427062376505 , 0.9106254691891366 , 0.2 )
( 0.8087453710717747 , 0.0873358066465128 , 0.8 )
( 0.22286168569889056 , 0.6672178044029601 , 0.2 )
( 0.22297990357259065 , 0.6156022713570786 , 0.2 )
( 0.8516233378780766 , 0.11998168903668985 , 0.8 )
( 0.19926435538276854 , 0.6994255744573358 , 0.2 )
( 0.38176637786578715 , 0.7412614969652668 , 0.2 )
( 0.12451618465828287 , 0.7369979876185243 , 0.2 )
( 0.10979748007965975 , 0.8512815509980239 , 0.2 )
( 0.10479225209789506 , 0.6579663171400711 , 0.2 )
( 0.7009834334711795 , 0.3720120811638443 , 0.8 )
( 0.7773961056233383 , 0.3599495053784583 , 0.8 )
( 0.8201340945592945 , 0.22151647062355037 , 0.8 )
( 0.131654833648978 , 0.816398412146849 , 0.2 )
( 0.6245513907080336 , 0.13009483281432893 , 0.8 )
( 0.430694727643888 , 0.7696278088773739 , 0.2 )
( 0.1972961888949818 , 0.7946443747181786 , 0.2 )
( 0.8941919835488764 , 0.3450755917066442 , 0.8 )
( 0.5897943690388352 , 0.26685768854268516 , 0.8 )
( 0.8482706859776603 , 0.404644785626021 , 0.8 )
( 0.4284911662411657 , 0.8111718300538617 , 0.2 )
( 0.4076415266589243 , 0.6543043974677796 , 0.2 )
( 0.16102372309131766 , 0.8368607468802273 , 0.2 )
( 0.2701167616942569 , 0.8142574315792738 , 0.2 )
( 0.1539149012676253 , 0.6483720921974706 , 0.2 )
( 0.2841292635125761 , 0.7305270744563124 , 0.2 )
( 0.9277902148119016 , 0.1938935051088681 , 0.8 )
( 0.27391219388005256 , 0.5725148987449746 , 0.2 )
( 0.36551326907490445 , 0.7713106951700408 , 0.2 )
( 0.0734335217243964 , 0.7396019784732524 , 0.2 )
( 0.6543978342157951 , 0.08650271025822998 , 0.8 )
( 0.6512898381036001 , 0.2000216613180242 , 0.8 )
( 0.1995156704890989 , 0.8120878635117923 , 0.2 )
( 0.821622311898793 , 0.2790205918520516 , 0.8 )
( 0.6962408637146551 , 0.08076108729329323 , 0.8 )
( 0.3636534200400908 , 0.8083595677085642 , 0.2 )
( 0.2679208190322526 , 0.6244303950353428 , 0.2 )
( 0.10389673801250013 , 0.8681040527778956 , 0.2 )
( 0.8033544609775752 , 0.36867725590955347 , 0.8 )
( 0.2740573751979177 , 0.7243578756697717 , 0.2 )
( 0.6948557533715101 , 0.18925959140395698 , 0.8 )
( 0.6728557525654665 , 0.2679928808749045 , 0.8 )
( 0.8647811959996955 , 0.18825570536856379 , 0.8 )
( 0.23518869693404476 , 0.9128898340829545 , 0.2 )
( 0.265926429611197 , 0.9133710202209234 , 0.2 )
( 0.6220106906020064 , 0.1219079342588435 , 0.8 )
( 0.6902126391538932 , 0.23480703283724697 , 0.8 )
( 0.08131589045249188 , 0.6954331462900474 , 0.2 )
( 0.7832485755015384 , 0.2709581697935482 , 0.8 )
( 0.19700724066228567 , 0.9067820388330438 , 0.2 )
( 0.7340980139455315 , 0.2627891475368584 , 0.8 )
( 0.22544026580990742 , 0.806029256851962 , 0.2 )
( 0.6648862020343914 , 0.3593651794811017 , 0.8 )
( 0.20270234187784264 , 0.640557176544345 , 0.2 )
       };
	\end{axis}
    	\end{scope}
        \draw[anchor=north] (0.8,-0.45) node {\tiny features + tag};
        \draw[anchor=north] (2.9,-0.45) node {\tiny distribution};
    	\end{scope}
	\end{scope}
\end{tikzpicture}
\end{center}
\vspace{-5mm}
\caption{\label{fig:histogram_spatial} \textbf{Spatial constraints via user-defined tags.}
The distribution of features does not encode their spatial organization.
In this example, \Hist cannot reproduce a checker pattern in image space (top).
Our trick consists in adding a spatial tag that acts like an additional dimension to the feature space. 
With this new dimension, the feature distribution can represent spatial structures that are processed by \Hist like any other feature space dimension (bottom).
}
\vspace{-5mm}
\end{figure}

\paragraph{Non-stationary statistics.}

Figure~\ref{fig:histogram_spatial}-top illustrates the case of feature activations that are organized like a checkerboard in image space.
Since the feature distribution is independent of the image-space locations of the features, optimizing the feature distribution does not preserve the checkerboard pattern.

\paragraph{Spatial constraints via user-defined tags.}

Our idea is to introduce spatial information in the feature distribution by adding a new dimension to the feature space that stores a spatial tag. 
In Figure~\ref{fig:histogram_spatial}-bottom, the 2D feature space becomes 3D and the third dimension stores a binary tag that encodes the checker 
pattern such that the only way for the optimized feature distribution to match the input distribution is to represent a checkerboard.  
This trick allows for adding spatial constraints that can be provided by user-defined spatial tags.
Note that the spatial tags are fixed, \ie they cannot be optimized, and they need to have the same distribution in the input and the output.

\paragraph{Homogeneity of the loss.}

An important point is to add the spatial dimension without breaking the homogeneity of the feature space loss \Hist. 
To do this, we concatenate spatial tags to the feature vectors $\left(F^l_m[1], .., F^l_m[N^l], \text{tag}\right)$, concatenate~1 to the normalized 
projection direction in feature space $\left(V_1, .., V_{N^l}, 1\right)$, and optimize for \Hist without further modifications. 
We use spatial tags that are strictly larger than the other dimensions in feature space such that the sorting in \Hist groups the 
pixels in clusters that have the same tag. 
As a result, the tags only change the sorting order and vanish after subtraction in Equation~\eqref{eq:slicing_loss}.
The introduction of the spatial dimension thus does not break the homogeneity of \Hist that remains a feature-space $L^2$ between sorted features.
\emph{No additional loss} and \emph{no meta-parameter tuning} is required with this approach.
Note that it is equivalent to solving separate $nD$ histogram losses for each cluster but it is more practical since it requires no more than a concatenation.

\paragraph{Results and positioning.}

\begin{figure}[!h]
\vspace{-3mm}
\begin{center}
\begin{tabular}{@{} c @{\hspace{1mm}} c @{}}
{\scriptsize\textbf{input + tag}} & 
{\scriptsize\textbf{optim} \textbf{with tag} {\small \Hist}} \\
\fbox{\includegraphics[width=0.49\linewidth]{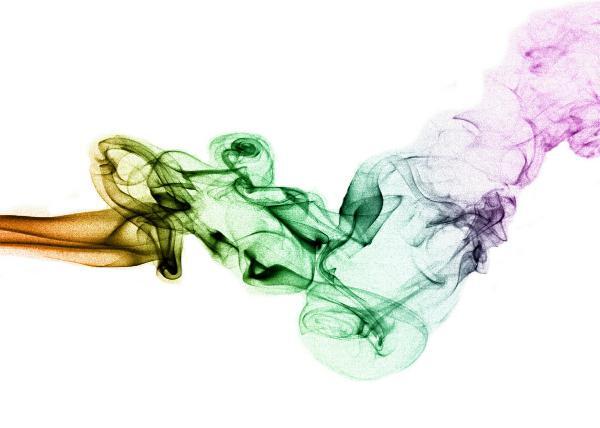}}
&
\fbox{\includegraphics[width=0.49\linewidth]{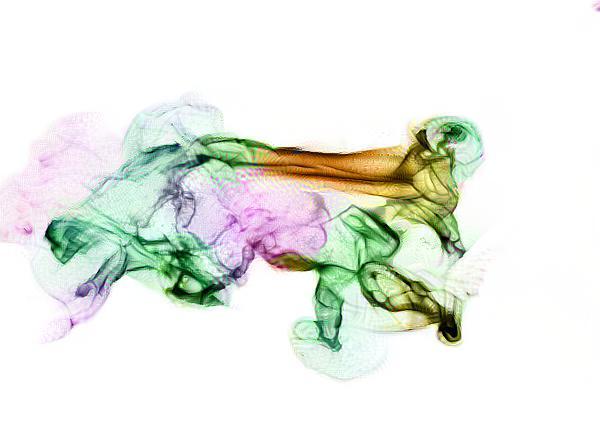}}
\\
\fbox{\includegraphics[width=0.49\linewidth]{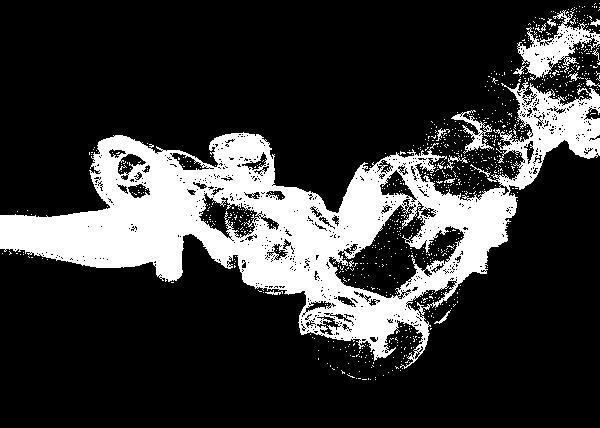}}
&
\fbox{\includegraphics[width=0.49\linewidth]{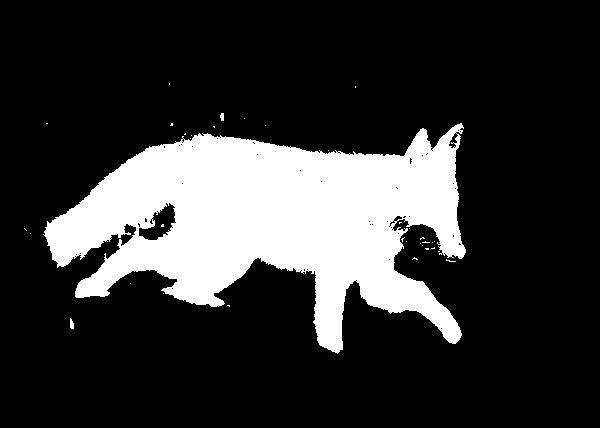}}
\\
\fbox{\includegraphics[width=0.49\linewidth]{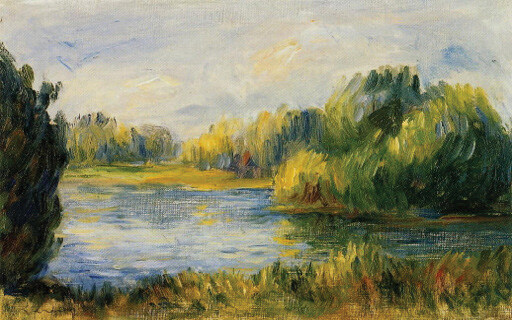}}
&
\fbox{\includegraphics[width=0.49\linewidth]{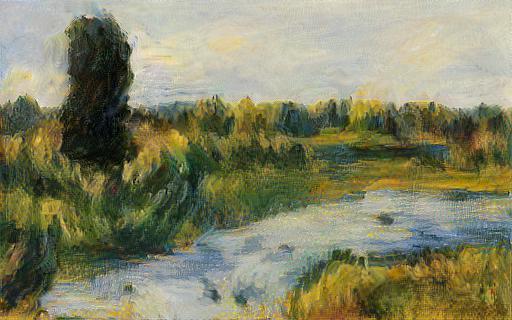}}
\\
\fbox{\includegraphics[width=0.49\linewidth]{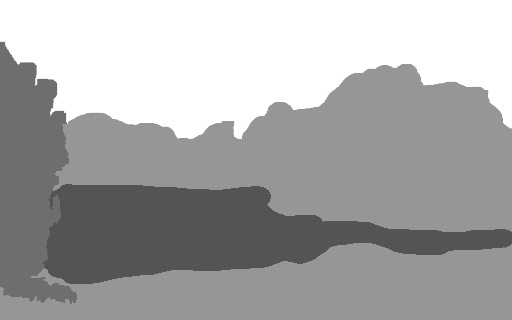}}
&
\fbox{\includegraphics[width=0.49\linewidth]{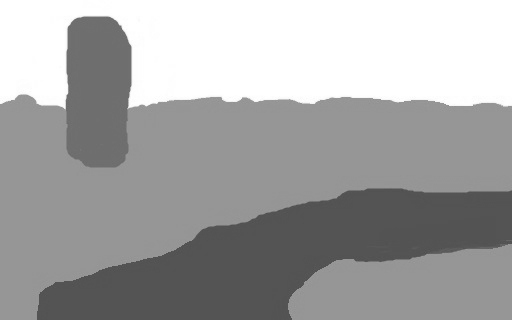}}
\end{tabular}
\end{center}
\vspace{-7mm}
\caption{\label{fig:texture_synthesis_spatial1} \textbf{Texture synthesis with spatial constraints: painting by texture.}
\Hist accounts for spatial tags concatenated to the deep features.
}
\vspace{-5mm}
\end{figure}

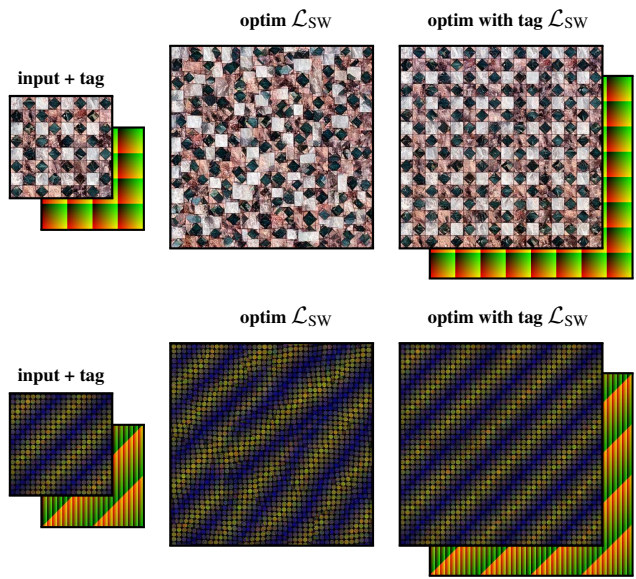
\begin{figure}[!h]
\begin{center}
\begin{tabular}{@{} c @{}}
\begin{tikzpicture}
\draw (0.0, 0.9) node[anchor=center] {\scriptsize \textbf{input + tag}};
\draw (2.85+0.15, 1.70) node[anchor=center] {\scriptsize\textbf{optim} {\small \Hist}};
\draw (5.95, 1.70) node[anchor=center] {\scriptsize\textbf{optim} \textbf{with tag} {\small \Hist}};
\draw (0.42, -0.42) node[anchor=center] {\fbox{\includegraphics[width=0.16\linewidth]{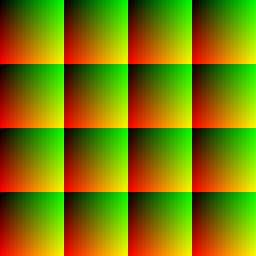}}};
\draw (0.0, 0.0) node[anchor=center] {\fbox{\includegraphics[width=0.16\linewidth]{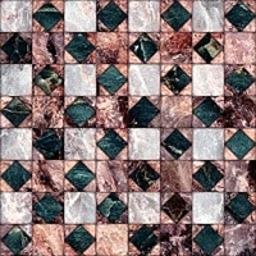}}};
\draw (2.65+0.15, 0.0) node[anchor=center] {\fbox{\includegraphics[width=0.32\linewidth]{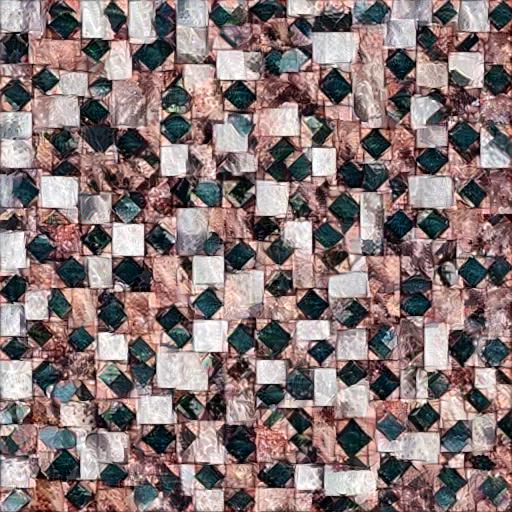}}};
\draw (6.25, -0.40) node[anchor=center] {\fbox{\includegraphics[width=0.32\linewidth]{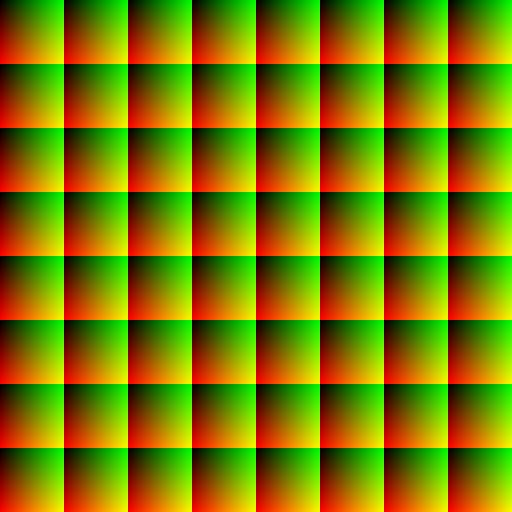}}};
\draw (5.85, 0.0) node[anchor=center] {\fbox{\includegraphics[width=0.32\linewidth]{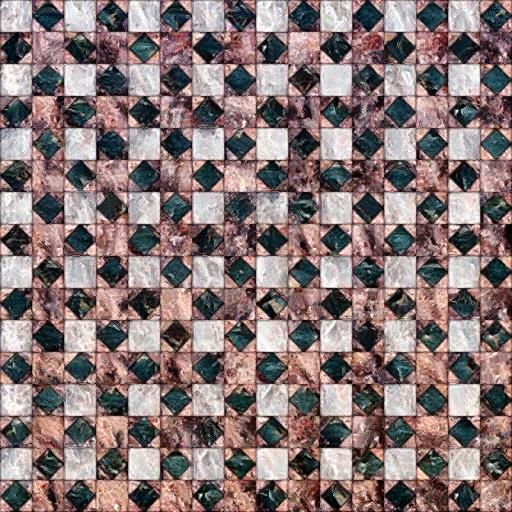}}};
\end{tikzpicture}
\\
\begin{tikzpicture}
\draw (0.00, 0.9) node[anchor=center] {\scriptsize \textbf{input + tag}};
\draw (2.85+0.15, 1.70) node[anchor=center] {\scriptsize\textbf{optim} {\small \Hist}};
\draw (5.95, 1.70) node[anchor=center] {\scriptsize\textbf{optim} \textbf{with tag} {\small \Hist}};
\draw (0.42, -0.42) node[anchor=center] {\fbox{\includegraphics[width=0.16\linewidth]{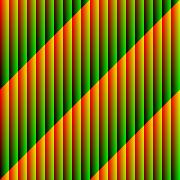}}};
\draw (0.0, 0.0) node[anchor=center] {\fbox{\includegraphics[width=0.16\linewidth]{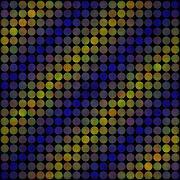}}};
\draw (2.65+0.15, 0.0) node[anchor=center] {\fbox{\includegraphics[width=0.32\linewidth]{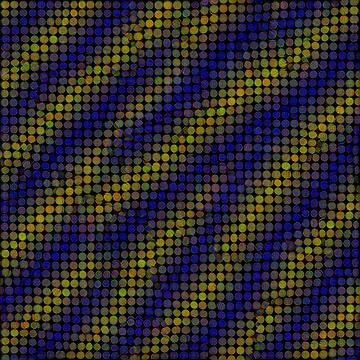}}};
\draw (6.25, -0.40) node[anchor=center] {\fbox{\includegraphics[width=0.32\linewidth]{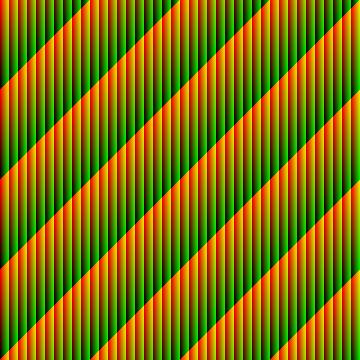}}};
\draw (5.85, 0.0) node[anchor=center] {\fbox{\includegraphics[width=0.32\linewidth]{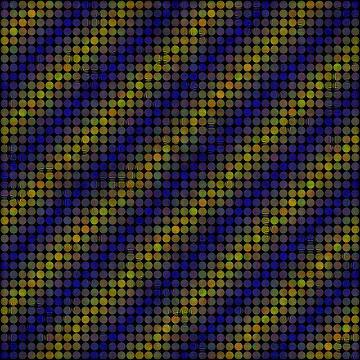}}};
\end{tikzpicture}
\end{tabular}
\end{center}
\vspace{-7.5mm}
\caption{\label{fig:texture_synthesis_spatial2} \textbf{Texture synthesis with spatial constraints: pseudo-periodic patterns.}
\Hist accounts for spatial tags concatenated to the deep features.
}
\vspace{-5mm}
\end{figure}

\begin{figure}[!h]
\begin{center}
\begin{tabular}{@{} c @{\hspace{1mm}} c @{\hspace{1mm}} c @{\hspace{1mm}} c @{}}
\textbf{Input} &
\textbf{Gatys~\etal}  &
\textbf{Sendik~\etal} &
\textbf{Ours}
\\
~ &
{\scriptsize \Gram} &
{\scriptsize $\alpha\mathcal{L}_{\scriptsize\text{Gram}} + \beta \mathcal{L}_{\scriptsize\text{DC}} +$} &
{\scriptsize \Hist with tag}
\\
~ &
 &
{\scriptsize $ \gamma \mathcal{L}_{\scriptsize\text{Div}} + \delta \mathcal{L}_{\scriptsize\text{S}}$} &

\\~\vspace{-2mm}\\
\fbox{\includegraphics[width=0.23\linewidth]{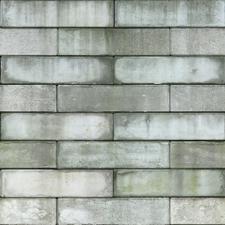}} &
\fbox{\includegraphics[width=0.23\linewidth]{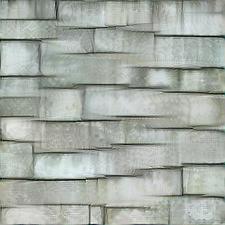}} &
\fbox{\includegraphics[width=0.23\linewidth]{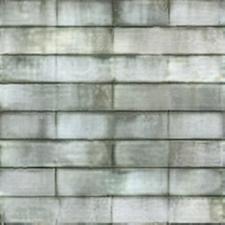}} &
 \fbox{\includegraphics[width=0.23\linewidth]{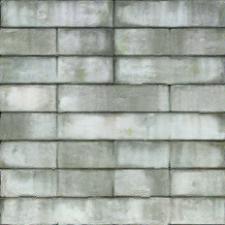}} 
\\
\fbox{\includegraphics[width=0.23\linewidth]{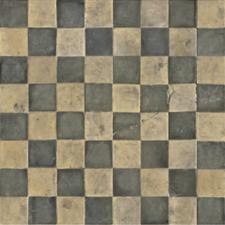}} &
\fbox{\includegraphics[width=0.23\linewidth]{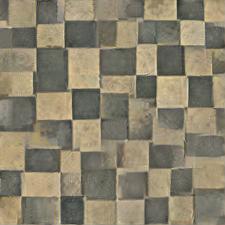}} &
\fbox{\includegraphics[width=0.23\linewidth]{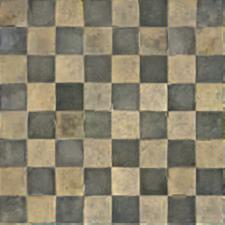}} & 
\fbox{\includegraphics[width=0.23\linewidth]{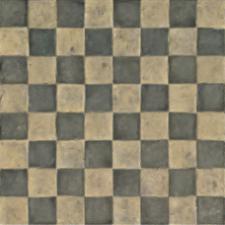}} 
\\
\fbox{\includegraphics[width=0.23\linewidth]{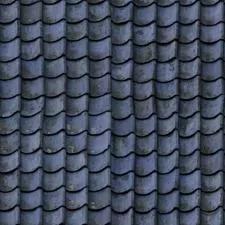}} &
\fbox{\includegraphics[width=0.23\linewidth]{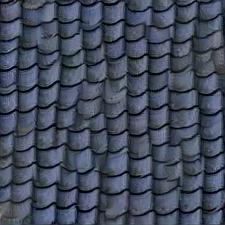}} &
\fbox{\includegraphics[width=0.23\linewidth]{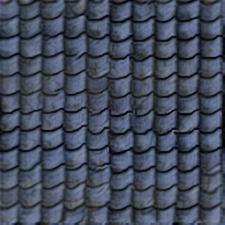}} & 
\fbox{\includegraphics[width=0.23\linewidth]{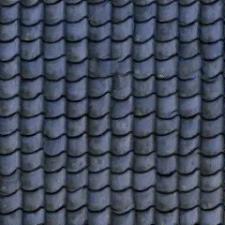}} 
\\
\fbox{\includegraphics[width=0.23\linewidth]{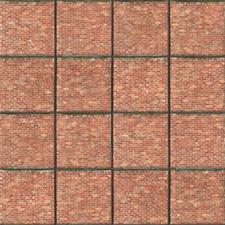}} &
\fbox{\includegraphics[width=0.23\linewidth]{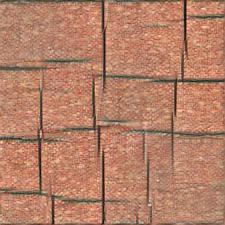}} &
\fbox{\includegraphics[width=0.23\linewidth]{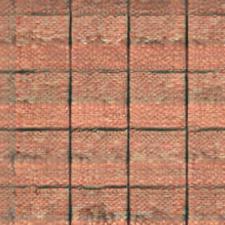}} & 
\fbox{\includegraphics[width=0.23\linewidth]{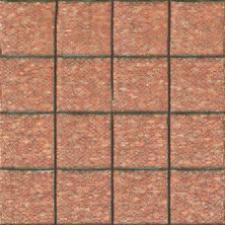}} 
\\
\fbox{\includegraphics[width=0.23\linewidth]{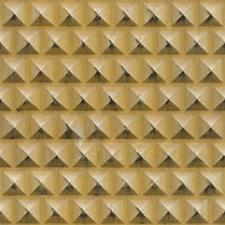}} &
\fbox{\includegraphics[width=0.23\linewidth]{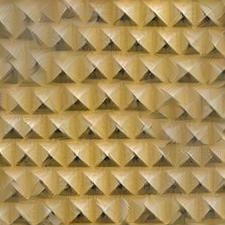}} &
\fbox{\includegraphics[width=0.23\linewidth]{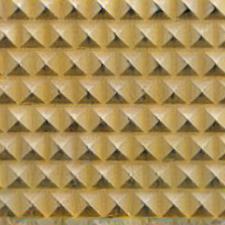}} & 
\fbox{\includegraphics[width=0.23\linewidth]{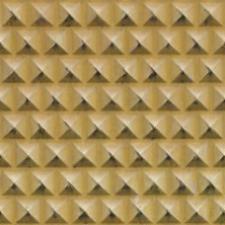}} 
\\
\fbox{\includegraphics[width=0.23\linewidth]{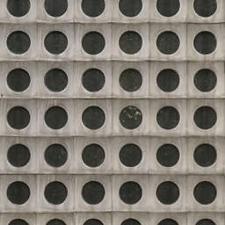}} &
\fbox{\includegraphics[width=0.23\linewidth]{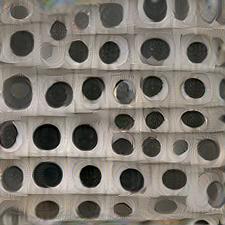}} &
\fbox{\includegraphics[width=0.23\linewidth]{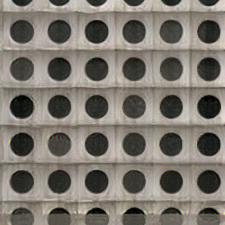}} & 
\fbox{\includegraphics[width=0.23\linewidth]{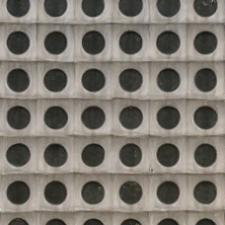}} 
\\
\fbox{\includegraphics[width=0.23\linewidth]{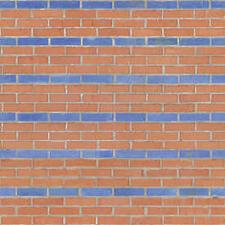}} &
\fbox{\includegraphics[width=0.23\linewidth]{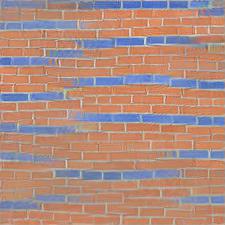}} &
\fbox{\includegraphics[width=0.23\linewidth]{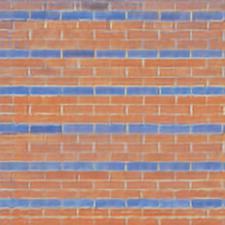}} & 
\fbox{\includegraphics[width=0.23\linewidth]{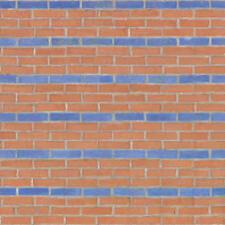}} 
\\
\fbox{\includegraphics[width=0.23\linewidth]{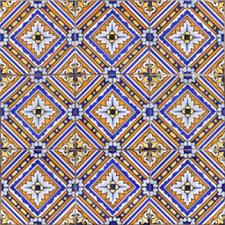}} &
\fbox{\includegraphics[width=0.23\linewidth]{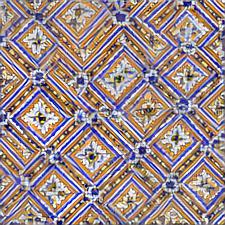}} &
\fbox{\includegraphics[width=0.23\linewidth]{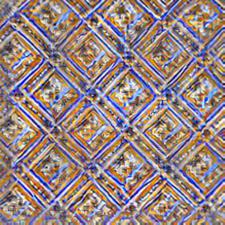}} & 
\fbox{\includegraphics[width=0.23\linewidth]{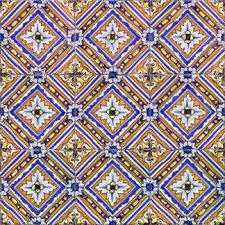}} 
\end{tabular}
\end{center}
\vspace{-6mm}
\caption{\textbf{Texture synthesis with spatial constraints: pseudo-periodic patterns.}
Comparison against Sendik and Cohen-Or's deep correlation method~\cite{DeepCorrelation}.
We extracted all the inputs and their results from their paper.
Our method uses spatial tags that encode the obvious pseudo-period such as in Figure~\ref{fig:texture_synthesis_spatial2}.
}
\label{fig:deepcorr}
\end{figure}

Figures~\ref{fig:texture_synthesis_spatial1}, \ref{fig:texture_synthesis_spatial2} and \ref{fig:deepcorr} show textures optimized with an L-BFGS-B optimizer and with spatial tags concatenated only to the first two layers of VGG-19.
Note that we do not aim at comparing the visual quality with competitor works.
Our point is to show that \Hist significantly widens the range of applications reachable with a \emph{single textural loss} and \emph{no meta-parameter tuning}.

\paragraph{Painting by texture.}

Figure~\ref{fig:texture_synthesis_spatial1} is an example of painting by texture where the user provides a target non-stationary example image, a spatial tag mask associated with the example, and a target spatial tag mask.
By optimizing \Hist with the spatial tags, we obtain a new image that has the same style as the example image but whose large-scale structure follows the desired tag mask. 
Typically, methods based on neural texture synthesis require parameter fine-tuning and/or an evaluation of the loss term \emph{for each tag} (cf. our discussion in Section~\ref{sec:previous}).
In comparison, \Hist works out of the box without any tuning and in a single evaluation regardless of the number of different of tags. 

\paragraph{Pseudo-periodic patterns.}

Figures~\ref{fig:texture_synthesis_spatial2} and \ref{fig:deepcorr} focus on textures with an obvious pseudo-period, which is provided as a spatial tag mask (the tag is the pixel coordinates modulo the period) and we optimize with \Hist. 
The results exhibit the structural regularity of the exemplar at that period, allowing for reproduction of regular textures with stochastic variation in all other frequencies.
In Figure~\ref{fig:deepcorr}, our results are comparable to Sendik and Cohen-Or~\cite{DeepCorrelation} for this class of textures.
Note that their method does not require user inputs but is significantly more elaborated and less efficient even for this simple class of textures. 
It uses \Gram in addition to three other losses whose weights need to be fine-tuned. 
Optimizing \Hist provides a simpler and efficient solution for this class of textures.

\clearpage 

\setlength{\fboxsep}{0pt}\setlength{\fboxrule}{0.8pt}

\setlength{\fboxsep}{0pt}\setlength{\fboxrule}{0.8pt}

\setlength{\fboxsep}{0pt}\setlength{\fboxrule}{0.8pt}

\clearpage


\section{Conclusion}
\label{sec:conclusion}

Our objective was to find a robust and high-quality textural loss. 
Several previous works show that the community felt that capturing the full feature distribution is the right approach for this problem.
However, existing approaches are subject to different shortcomings such as quadratic complexity, additional regularization terms, etc. 
Surprisingly, there existed a much simpler solution from the start in the optimal transport community.
The Sliced Wasserstein Distance provides a textural loss with \textit{proven convergence}, \textit{sub-quadratic complexity}, \textit{simple implementation} and that achieves \textit{high-quality} \textit{without further regularization losses}.
It seems to be the right tool for this problem and we are not aware of any alternative that brings all these qualities together.
We benchmarked and validated it in a texture optimization framework, which we believe is the correct way to unit test and validate a textural loss: the results reflect the performance and expressiveness of the loss without being hindered by a generative architecture. 
Nonetheless, we have also shown that one can successfully use it to train a generative architecture. 
Finally, we have shown a simple way to handle spatial constraints that widens the expressiveness of the loss without compromising its simplicity, which is, to our knowledge, not possible with the Gram-matrix loss and other alternatives.  
With these good properties, we hope that the Sliced Wasserstein Distance will be considered as a serious competitor for the Gram-matrix loss in terms of practical adoption for both education, research and production.

\appendix

\section{On Texture Synthesis vs. Style Transfer}
\label{sec:appendix}

In this section, we show that style-transfer methods should not be expected to perform well on texture synthesis unless they are proven to do so. 
For instance, the textural losses \Context by Mechrez~\etal~\cite{ContextualLoss} and \STROTSS by Kolkin~\etal~\cite{Kolkin_2019_CVPR} are state-of-the-art for style transfer but, in our experience, they perform worse than a vanilla \Gram when used for pure texture synthesis.
\textbf{The right ablation study to evaluate texture quality is texture synthesis, not style transfer in which other effects are mixed in}. 
To support this point, we took the implementations of \Context and \STROTSS provided by their authors and adapted them to texture synthesis. 

\begin{figure}[!h]
    \begin{center}
        \begin{tabular}{@{\hspace{0mm}} c @{\hspace{0.5mm}} c @{\hspace{0.5mm}} c @{}}
             \textbf{Input} 
		 &  \Gram		 
            &  \Context		 
		  \\
		\scalebox{0.5}{$100\times100$} &
		\scalebox{0.5}{$100\times100$} &
		\scalebox{0.5}{$100\times100$} 
		\\
            \fbox{\includegraphics[width=.32\linewidth]{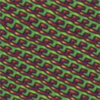}}
		 & \fbox{\includegraphics[width=.32\linewidth]{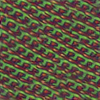}}
            & \fbox{\includegraphics[width=.32\linewidth]{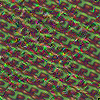}} 		 
		 \\
            \fbox{\includegraphics[width=.32\linewidth]{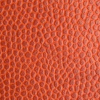}}       
		 & \fbox{\includegraphics[width=.32\linewidth]{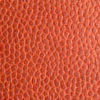}}        
            & \fbox{\includegraphics[width=.32\linewidth]{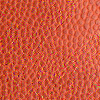}}   		  
        \end{tabular}
    \end{center}
    \vspace{-6mm}
    \caption{\textbf{Texture optimization using \Context~\cite{ContextualLoss}.} 
    Results are limited to a resolution of $100\times 100$ and are of low quality.
    }
    \label{fig:context}
\vspace{-6mm}
\end{figure}

In Figure~\ref{fig:context}, we show a texture synthesis experiment with the implementation of \Context provided by Mechrez~\etal~\cite{ContextualLoss} (in which we disabled the content loss).
The memory allocator crashes beyond a resolution of $100^2$ due to $\mathcal{O}(n^2)$ complexity (they limit to $65^2$) and the results are visually worse than with \Gram (we tried several values for their $h$ parameter and kept the best results).

\begin{figure}[!h]
    \begin{center}
\begin{tabular}{@{} c @{\hspace{0.5mm}} c @{}}
{\textbf{Input}} &
{\Gram}
\vspace{-1mm} \\ 
\scalebox{0.5}{$256\times256$} &
\scalebox{0.5}{$512\times512$} \\
\raisebox{8mm}{\fbox{\includegraphics[width=0.22\linewidth]{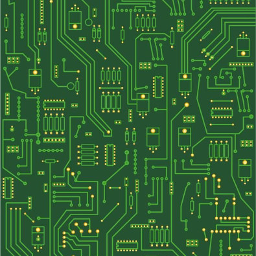}}} &
\fbox{\includegraphics[width=0.44\linewidth]{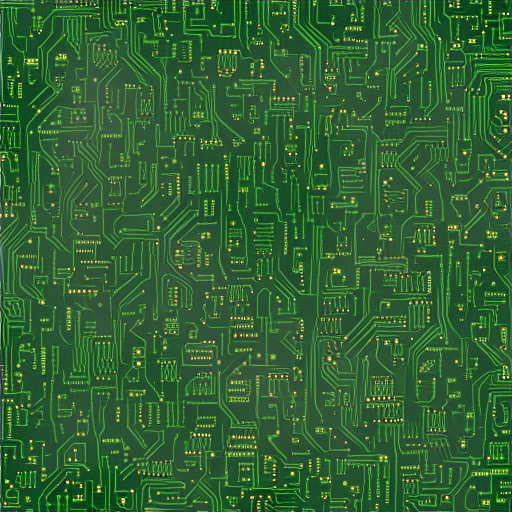}}  \\ 
\STROTSSwithreg&
\STROTSSnoreg  
\vspace{-1mm} \\ 
\scalebox{0.5}{$512\times512$} &
\scalebox{0.5}{$512\times512$} 
\\
\fbox{\includegraphics[width=0.44\linewidth]{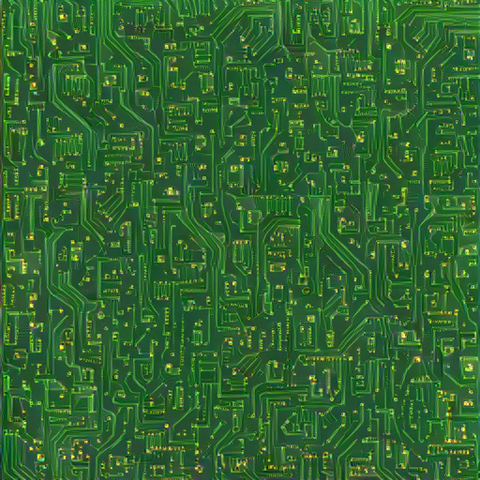}} &
\fbox{\includegraphics[width=0.44\linewidth]{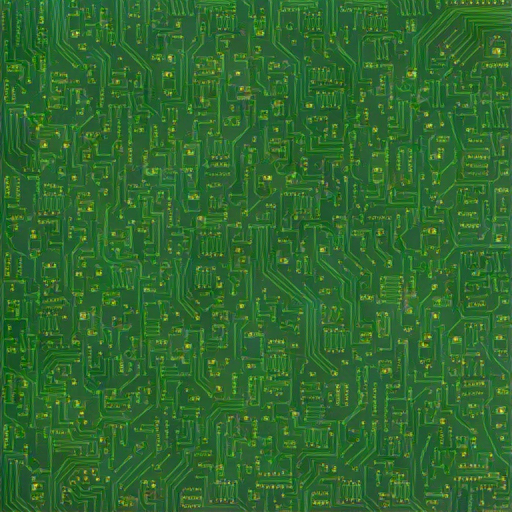}} 
\end{tabular}
    \end{center}
    \vspace{-6mm}
\caption{
\textbf{Texture optimization using \STROTSS~\cite{Kolkin_2019_CVPR}.}
The loss achieves poor textural quality by itself. 
The textural quality remains inferior to \Gram even when using the proposed additional regularization losses $\mathcal{L}_\text{m}$ and  $\mathcal{L}_\text{p}$.
}
\label{fig:strotss}
\vspace{-6mm}
\end{figure}

In Figure~\ref{fig:strotss}, we show a texture synthesis experiment with the implementation of \STROTSS provided by Kolkin~\etal~\cite{Kolkin_2019_CVPR} on a texture extracted from their paper (again, we disabled content loss). 
The result is qualitatively worse than with \Gram.
Furthermore, \STROTSS needs two additional regularization losses ($\mathcal{L}_\text{m}$ and  $\mathcal{L}_\text{p}$) without which quality decreases significantly.

From these experiments, we conclude that \Context and \STROTSS are not good candidates for texture synthesis and that, more generally, \textbf{state-of-the-art style transfer methods are not necessarily good candidates for texture synthesis}.

\setlength{\fboxsep}{0pt}\setlength{\fboxrule}{0.8pt}

\setlength{\fboxsep}{0pt}\setlength{\fboxrule}{0.8pt}

{\small
\bibliographystyle{ieee_fullname}
\bibliography{papers}
}

\end{document}